\PassOptionsToPackage{colorlinks=true}{hyperref}
\documentclass[lettersize,journal]{IEEEtran}
\usepackage{amsmath,amsfonts}
\usepackage{array}
\usepackage[caption=false,font=normalsize,labelfont=sf,textfont=sf]{subfig}
\usepackage{textcomp}
\usepackage{stfloats}
\usepackage{verbatim}
\usepackage{graphicx}
\usepackage{cite}  
\usepackage{multirow}
\usepackage{lineno}
\usepackage{makecell}
\usepackage{amssymb}
\usepackage{url}
\usepackage[table]{xcolor} 
\usepackage{pifont}
\usepackage{listings}
\usepackage{booktabs}
\usepackage{orcidlink}
\usepackage[ruled,boxed,linesnumbered]{algorithm2e}

\usepackage{hyperref}
\usepackage{cleveref}

\crefname{section}{Sec.}{Secs.}
\Crefname{section}{Sec.}{Secs.}
\crefname{subsection}{Sec.}{Secs.}
\Crefname{subsection}{Sec.}{Secs.}
\crefname{subsubsection}{Sec.}{Secs.}
\Crefname{subsubsection}{Sec.}{Secs.}

\crefname{table}{Tab.}{Tabs.}
\Crefname{table}{Tab.}{Tabs.}

\crefname{figure}{Fig.}{Figs.}
\Crefname{figure}{Fig.}{Figs.}

\Crefname{equation}{Eq.}{Eq.}
\crefname{equation}{Eq.}{Eq.}

\Crefname{algorithm}{Alg.}{Alg.}
\crefname{algorithm}{Alg.}{Alg.}

\definecolor{codeblue}{rgb}{0.25,0.5,0.75}
\definecolor{codegray}{rgb}{0.5,0.5,0.5}
\definecolor{codepurple}{rgb}{0.58,0,0.82}
\definecolor{codepink}{rgb}{0.8, 0, 0.45}

\lstdefinestyle{mypython}{
  language=Python,
  basicstyle=\ttfamily\small,
  keywordstyle=\color{codepink}\bfseries,
  commentstyle=\color{codegray}\itshape,
  stringstyle=\color{codepurple},
  showstringspaces=false,
  breaklines=true,
  tabsize=2,
  frame=none,
  numbers=left,
  numberstyle=\small\color{codegray},
  numbersep=6pt,
  xleftmargin=-1pt,
  alsoletter=+-*/=<>!,
  moredelim=[is][\color{codepink}]{@@}{@@},
  literate=
    {.}{{\textbf{.}}}1
    {+}{{{\color{codepink}+}}}1
    {-}{{{\color{codepink}-}}}1
    {*}{{{\color{codepink}*}}}1
    {/}{{{\color{codepink}/}}}1
    {=}{{{\color{codepink}=}}}1
    {>}{{{\color{codepink}>}}}1
    {<}{{{\color{codepink}<}}}1
    {!}{{{\color{codepink}!}}}1
    {:}{{{\color{codepink}:}}}1
}

\lstnewenvironment{PythonB}[1][]{
    \lstset{style=mypython,#1}
}{}

\def\ie{\textit{i.e.}~}
\def\wrt{\textit{w.r.t.}~}
\def\eg{\textit{e.g.}~}
\def\etc{\textit{etc.}}

\def\intok{\textbf{INT}\xspace}

\def\ourmodel{\textbf{EI}\xspace}

\newcommand{\xmark}{\ding{55}}%

\begin{document}

\title{Early Intervention for VFM-based Multimodal Medical Image Classification}

\author{Qijie Wei\orcidlink{0009-0008-6895-5870}, Hailan Lin\orcidlink{0009-0004-0529-3196}, and Xirong Li\orcidlink{0000-0002-0220-8310}, \IEEEmembership{Member, IEEE}
\thanks{This work was funded by National Natural Science Foundation of China (62576348) and
Beijing Natural Science Foundation (L254039). \emph{(Corresponding author: Xirong Li.)}}
\thanks{
Qijie Wei, Hailan Lin, and Xirong Li are with Renmin University of China, Beijing 100872, China (e-mail: \{qijie.wei, hailan\_lin, xirong\}@ruc.edu.cn)}
}


\maketitle

\begin{abstract}

Current methods for multimodal medical image classification (M3IC) face two major challenges. First, the prevailing ``fusion after unimodal image embedding'' paradigm cannot fully exploit the complementary and correlated information in the multimodal data. Second, the scarcity of labeled multimodal medical images, coupled with their substantial domain shift from natural images, impedes the use of cutting-edge Vision Foundation Models (VFMs) for medical image embedding. To jointly address the challenges, we propose a novel \emph{\textbf{Early Intervention}} (\ourmodel) framework. Treating one modality as target and the rest as reference, \ourmodel harnesses high-level semantic tokens from the reference as \textbf{intervention tokens} to steer the target modality's embedding process at an early stage. For parameter-efficient VFM adaptation, we introduce \textbf{M}ixture \textbf{o}f varied-\textbf{r}ank LoRAs (\textbf{MoR}), which employs a small set of LoRA experts with distinct ranks and a bypass-allowed router. Extensive experiments on three public datasets, covering retinal disease recognition, skin lesion recognition, and knee anomaly classification, verify the effectiveness of the proposed method against a number of competitive baselines.

\end{abstract}


\section{Introduction}\label{sec:intro}

\IEEEPARstart{M}ultimodal medical imaging is vital in modern healthcare, providing objective clinical profiles by visualizing organs through complementary imaging devices and viewpoints \cite{kumar2024multimodality}. Recognizing lesions or diseases from such multimodal image data, \ie \underline{m}ulti\underline{m}odal \underline{m}edical \underline{i}mage \underline{c}lassification (M3IC), is therefore an important task in medical computer vision. In this paper, we tackle the M3IC task with a novel Vision Foundation Model (VFM) based method.

Each modality in a multimodal image sample provides unique and complementary information. In fundus imaging, for instance, color fundus photography (CFP) captures the \emph{en face} of the retina, showing vessels, the optic disc, and the macula, while optical coherence tomography (OCT) reveals the retina’s cross-sectional structure \cite{mm-mil}. How to effectively extract and fuse discriminative features from the multimodal sample is crucial for generating embeddings useful for retinal disease recognition.

\begin{figure}[!tbp]
  \centering
  \subfloat[Previous: Cross-modal guidance \emph{after} UIE]{
    \includegraphics[width=0.9\columnwidth]{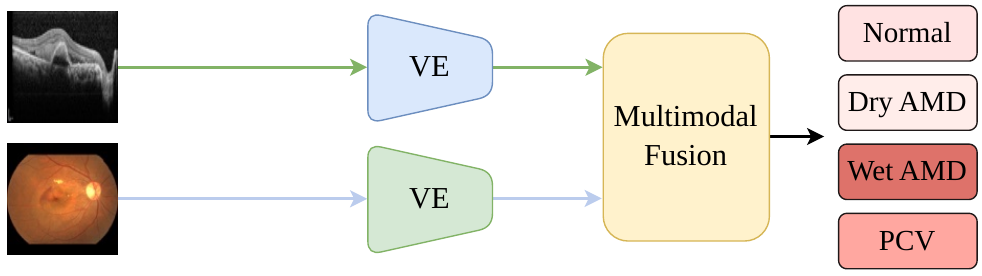}
    \label{fig:sketch-a}
  }

  \subfloat[Proposed: Cross-modal guidance \emph{before} UIE]{
    \includegraphics[width=0.9\columnwidth]{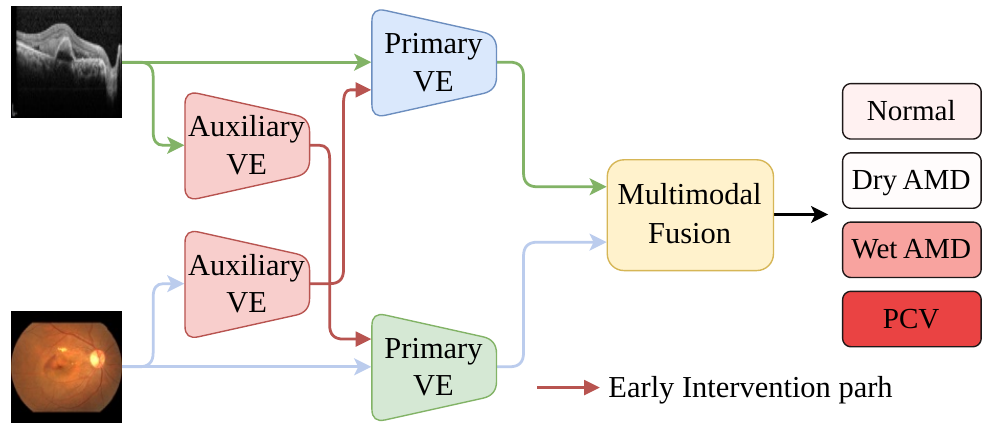}
    \label{fig:sketch-b}
  }
  \caption{\textbf{High-level diagram of the proposed early intervention (\ourmodel) framework for multimodal medical image classification (M3IC)}. Different from the prevailing ``fusion after unimodal image embedding'' (UIE) paradigm, \ourmodel is the first to enable cross-modal guidance before UIE. Vision encoders (VE) previously used solely for UIE now serve two distinct roles: primary and auxiliary. With each modality treated as the target and the rest as its reference, the auxiliary VE extracts intervention (\intok) tokens from the reference. These \intok tokens are then merged with low-level tokens from the target modality, steering the primary VE's embedding process at an early stage.}
  
\end{figure}

Existing methods, \eg MM-MIL \cite{mm-mil} for retinal disease recognition, CosCatNet \cite{CosCatNet} for skin lesion classification and RadDiag \cite{ RadDiag} for radiology-based disease diagnosis, generate the embeddings in a two-step manner, see \cref{fig:sketch-a}. Firstly, unimodal image embedding (UIE) is performed with modality-specific vision encoders (VEs) to extract features per modality. These unimodal features are then fused via vector concatenation \cite{CosCatNet}, weighted summation \cite{mm-mil} or self attention \cite{RadDiag} to produce a joint representation. Despite their varied fusion strategies, these methods have a common pattern:  \textbf{fusion after UIE}, see \cref{tab:related_work}.

We argue that introducing cross-modal guidance after UIE is too late to fully leverage the complementary and correlated information in the multimodal data. In fact, such a \emph{delayed} inter-modal interaction is also misaligned with clinical practice, where a clinician rarely interprets modalities in isolation. Instead, they often form a first impression from one modality and use it to contextually guide the interpretation of another. We identify the delayed fusion paradigm as a key bottleneck that hinders the advancement of M3IC.

\begin{table}[!htbp]
\caption{\textbf{Current methods for M3IC}. 
Different from prior works, we propose \ourmodel for cross-modal guidance \emph{before} UIE and Mixture of varied-rank LoRAs (\emph{MoR}) for parameter-efficient fine-tuning (PEFT) of vision foundation models (VFMs). }
\label{tab:related_work}
\renewcommand{\arraystretch}{1.2}
\resizebox{\linewidth}{!}{
\begin{tabular}{@{}lllll@{}}

\toprule
\textbf{Method} &  \textbf{UIE}  & \textbf{\makecell{Cross-modal\\guidance}} & \textbf{PEFT} & \textbf{M3IC Tasks}\\
\hline
MM-MIL\cite{mm-mil} & ResNet & After UIE & \xmark & $\cdot$Retinal disease classification\\
\rowcolor{gray!20}
CosCatNet\cite{CosCatNet} & ResNet  & After UIE & \xmark & $\cdot$Skin lesion recognition  \\
RadDiag\cite{RadDiag} & ResNet+ViT & After UIE & \xmark & $\cdot$Radiology based diagnosis \\
\rowcolor{gray!20}
GeCoM-Net\cite{GeCoM-Net} & DenseNet  & After UIE & \xmark & $\cdot$Retinal disease classification \\
MMRAD\cite{mmrad} & SAM & After UIE & Prompt & $\cdot$Retinal anomaly detection \\
\rowcolor{blue!20}
\ourmodel & \makecell[l]{CLIP,\\ DINOv2} & \emph{Before} UIE & MoR & \makecell[l]{$\cdot$Retinal disease classification\\$\cdot$Skin lesion recognition\\$\cdot$Knee anomaly classification} \\
\bottomrule

\end{tabular}}
\end{table}

%

For learning effective multimodal medical image embeddings, innovating the fusion paradigm alone is insufficient. Given the scarcity of labeled data in the medical domain, data-efficient learning algorithms also matter. VFMs like CLIP \cite{clip} and DINOv2, pre-trained on massive amounts of Internet images, provide powerful representations for natural images. However, the significant domain gap between natural and medical images makes their direct application problematic. Current methods for M3IC mostly rely on full-parameter tuning of ImageNet-pretrained encoders (\cref{tab:related_work}). 
MMRAD \cite{mmrad} makes an initial attempt to adapt a VFM using prompt learning. This technique aims to activate a model's existing task-specific knowledge by prepending a set of learnable prompts to the input, see VPT \cite{vpt} for generic image classification, DVPT \cite{dvpt}, FPT+ \cite{fptp} and ScaPT \cite{dong2024prompt} for \emph{unimodal} medical image classification. Despite its data efficiency, prompt learning becomes ineffective when the model intrinsically lacks  knowledge \wrt the task of interest. 

To effectively adapt a given VFM to the medical domain where labeled samples are in relatively short supply,  a growing line of research is to insert small trainable modules, known as adapters, into a frozen backbone. Examples include LKA \cite{lka} and FreqFiT \cite{freqfit} for medical image classification and LateLoRA \cite{latelora} for medical image segmentation. Nonetheless, these methods are all designed for unimodal scenarios. To the best of our knowledge, how to best leverage cutting-edge VFMs for M3IC remains largely unexplored.

\begin{figure}[!htbp]
  \centering
    \subfloat[Multimodal image pairs of retina (MMC-AMD \cite{jbhi22-mmcamd})]{
    \includegraphics[width=\columnwidth]{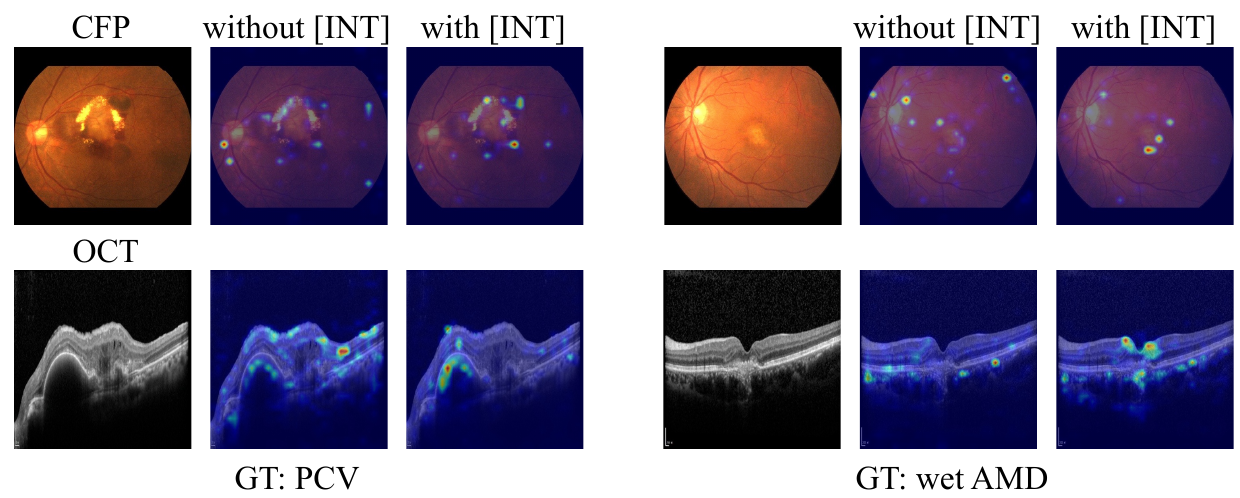}
    \label{fig:att1}
  }

  \subfloat[Multimodal image pairs of skin (Derm7pt \cite{derm7pt})]{
    \includegraphics[width=\columnwidth]{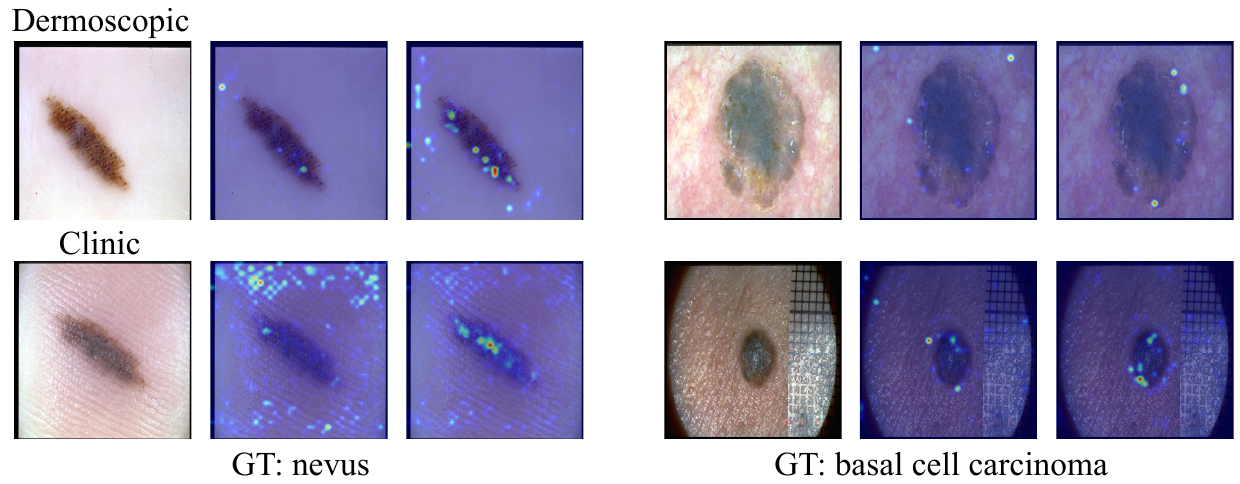}
    \label{fig:att2}
  }

  \subfloat[Multi-view MRI images of knee (MRNet \cite{mrnet})]{
    \includegraphics[width=\columnwidth]{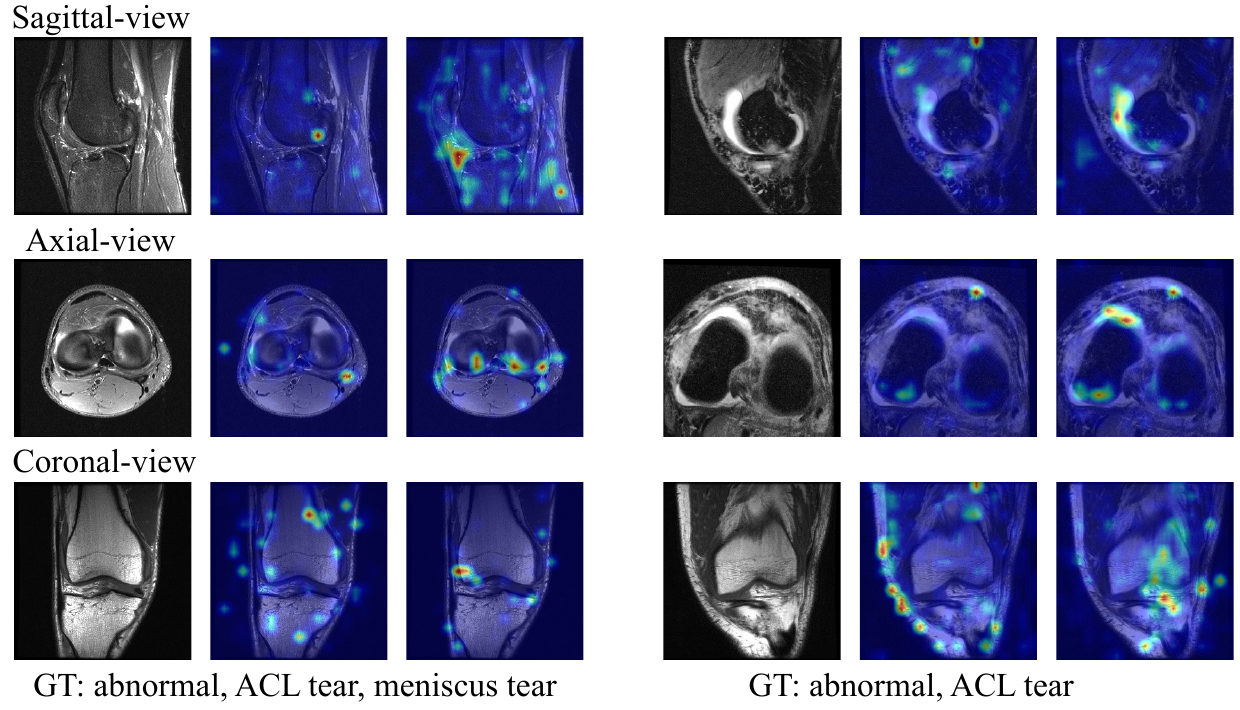}
    \label{fig:att3}
  }
  \caption{\textbf{Multimodal medical images and their patch-level similarity maps \wrt the CLS token}. VFM: DINOv2. As CLS is used for classification, such maps reflect patch-wise contributions to the final prediction. Per target modality (say CFP), the inclusion of the \intok tokens from its reference modality (say OCT) leads to more \emph{lesion}-focused maps. Best viewed in color.}
  \label{fig:att}
\end{figure}

In this paper, we propose \textbf{Early Intervention} (\ourmodel), a new framework for M3IC that introduces cross-modal guidance \emph{before} UIE, see \cref{fig:sketch-b}. Viewing one modality as target and the rest as reference, \ourmodel leverages the high-level semantics from reference modalities, extracted by auxiliary VFMs, as \textbf{intervention tokens} (\intok) to guide UIE of the target modality at a very early stage. This design mirrors a clinician’s typical workflow, where an initial interpretation or first impression of one modality contextually informs the examination of another. As demonstrated in \cref{fig:att}, the inclusion of the \intok tokens leads to more \emph{lesion}-focused feature extraction.
Moreover, inspired by the success of LoRA \cite{lora}, 
we introduce \textbf{M}ixture \textbf{o}f varied-\textbf{r}ank LoRAs (MoR), which employs a set of LoRA experts with different ranks and  a \emph{bypass}-allowed router 
%
%
for more effective and parameter-efficient VFM adaptation.

In sum, our work makes three contributions as follows: 
\begin{itemize}
    \item \textbf{A novel multimodal fusion framework}. The proposed early intervention path effectively propagates cross-modal guidance (\cref{fig:architecture}), thereby overcoming the information bottleneck in previous methods caused by the conventional fusion-after-UIE mechanism.
    \item \textbf{A better PEFT method}. Compared to its predecessors (LoRA,  LoRAMoE \cite{lora-moe}, AdaLoRA \cite{adalora}, FlyLoRA \cite{flylora}, \etc), the proposed MoR uses a parameter-efficient router (PE-Router) to exploit multiple LoRA experts of distinct ranks in an input-dependent manner. Moreover, the PE-Router contains a \emph{bypass} pathway that can skip these experts when the original weights suffice. MoR enables more effective adaptation of VFMs to medical images.
    \item \textbf{State-of-the-art performance}. We conduct extensive experiments on three M3IC datasets, \ie MMC-AMD \cite{jbhi22-mmcamd}, Derm7pt \cite{derm7pt} and MRNet \cite{mrnet}, for fundus, skin and knee image classification, respectively. The results consistently show the superiority of the proposed solution over a number of competitive baselines. \ourmodel is open source\footnote{\url{https://github.com/ruc-aimc-lab/EI}}.
\end{itemize}

A preliminary version of this work appeared in the CVPR 2026 Findings \cite{EI}. 
Compared to the conference paper, the journal version makes multiple technical improvements as follows. First, better \intok tokens. For each reference modality, we enrich its \intok tokens by a small set of contextual tokens, incurring negligible computational overhead. Second, parameter-efficient router. Given an MoR comprising three LoRA experts, our $\ell_2$-norm based routing strategy reduces the number of the router parameters by 75\%. Third, better initialization of low-rank factors. For each expert, we provide a \emph{near}-zero start with \emph{nonzero} low-rank factors, enabling effective adaptation learning while minimally perturbing the pre-trained VFM when the learning starts. These improvements naturally lead to substantially extended experiments.

\section{Related Work} \label{sec:related}

Our work sits at the crossroads of embedding-based M3IC and LoRA-based PEFT. We briefly review recent developments in both areas and, accordingly, clarify our novelties. Since M3IC is essentially a multimodal fusion problem, we broaden our discussion to encompass multimodal fusion techniques that can be readily adapted to M3IC.

\subsection{Embedding-based Methods for M3IC} \label{ssec:related-img-embedding}

Despite its importance, M3IC is surprisingly understudied. Given a multimodal medical image sample, depending on how its modality-specific embeddings are combined,  we categorize existing methods into three groups:  concatenation \cite{CosCatNet,GeCoM-Net}, weighted summation \cite{mm-mil} and attention-based fusion \cite{RadDiag,mmrad}.


For skin lesion classification, CosCatNet  first uses two ResNets to obtain dermoscopic and clinic image features and then concatenates them to obtain a multimodal embedding \cite{CosCatNet}. The concatenation operation is also used in GeCoM-Net to fuse CFP and OCT features for multimodal retinal disease recognition \cite{GeCoM-Net}. In a similar context, MM-MIL first uses two ResNets in parallel to extract features from given CFP and OCT images, respectively \cite{mm-mil}. The resultant features are then weightedly summed up, with the weights determined by a lightweight multiple instance learning module.
Regarding attention-based fusion, RadDiag  employs self-attention  to integrate embeddings of multimodal radiological images for radiology-based long-tailed disease diagnosis \cite{RadDiag}.
MMRAD instead uses two cross-attention blocks in sequence to fuse embeddings of CFP and OCT images for retinal anomaly detection \cite{mmrad}. 
Despite their varied fusion strategies, the above methods share a common pattern: fusion after UIE, see \cref{tab:related_work}.
UIE is a crucial step, yet uninformed by other modalities. The proposed \ourmodel framework addresses this fundamental issue by enabling cross-modal guidance \emph{before} UIE.



\subsection{LoRA-based Methods for PEFT} \label{ssec:related-peft}

Since the introduction of LoRA \cite{lora}, LoRA-based methods have been a prominent approach to PEFT. LoRA is typically applied to the \texttt{linear} layers of a given foundation model. Specifically, consider a linear layer with a frozen weight matrix $W \in \mathbb{R}^{d \times d}$. Let $h$ and $h'$ denote the layer input and output, respectively, both in $\mathbb{R}^{d}$. A LoRA module with rank $r$ (typically 4) is parameterized by two low-rank factors $A \in \mathbb{R}^{r \times d}$ and $B \in \mathbb{R}^{d \times r}$, and its output is given by $B A h$. By adding the LoRA output to the original output of the linear layer, an updated version of  $h'$ is obtained as
\begin{equation} \label{eq:lora}
h' = Wh + \underbrace{BA}_{\Delta W}h,
\end{equation}
where $\Delta W=BA$ denotes the adapter weights, and $\Delta W=0$ means using the original weights without adaptation.

Subsequent works have improved LoRA primarily in three aspects: enhanced network structure \cite{losa,lora-moe}, task-adaptive rank selection \cite{adalora,sora,flylora} and better initialization of the  low-rank factors \cite{li2025beyond,meng2024pissa,wang2024loraga}.
In the first aspect, LoSA attaches auxiliary adapters to the backbone for more efficient gradient propagation \cite{losa}. LoRAMoE employs a mixture of LoRAs per linear layer \cite{lora-moe}.  These works use a fixed rank of 4, following the original LoRA. However, we empirically observe that the optimal rank depends on both the task and the individual input.

As for the second aspect, 
AdaLoRA zeros out less important components of the low-rank factors in a layer-wise manner \cite{adalora}. 
SoRA controls the effective rank by applying a learnable sparsity gate to the low-rank factors \cite{sora}.
Note that in both methods, the rank is fixed once selected and input-independent during inference. 
FlyLoRA measures the importance of each of the $r$ vectors in $A$ \wrt the input $h$ using the absolute value of the corresponding element in the down-projected vector $Ah$ \cite{flylora}. In particular, the top-$k$ elements with the largest absolute values are retained, with the rest zeroed out. This design enables FlyLoRA to selectively activate $k$ vectors in $A$ per input. However, the value of $k$ remains fixed.




Concerning the initialization of the low-rank factors, the original implementation \cite{lora} uses a random Gaussian matrix for $A$, zero for $B$, and thus zero for $\Delta W$. While this ensures zero adaptation at the beginning of training, it introduces the gradient vanishing issue to $A$ \cite{hayou2024impact,li2025beyond}. To tackle the issue and reduce the number of trainable parameters, UORA \cite{uora} and RandLoRA \cite{randlora} initialize both $A$ and $B$ as products of a random matrix and a diagonal matrix, and train only the diagonal matrices. In contrast to random initialization, LoLDU obtains low-rank factors by performing an LDU decomposition of $W$ \cite{loldu}, and trains only the diagonal component of the decomposition. PiSSA initializes $A$ and $B$ with the top singular components of $W$ to let $A$ span the most informative directions of the original weights \cite{meng2024pissa}. Alternatively, LoRA-GA performs SVD on the gradients of $W$ to achieve gradient-wise alignment between $\Delta W$ and $W$  \cite{wang2024loraga}. As the above methods consider only the single-LoRA setting, they are suboptimal for the multi-LoRA setting explored in this work.

Our proposed MoR improves LoRA across all three aspects. It employs a mixture of LoRAs with a bypass-allowed router to adaptively accept or reject  adaptation per input. The LoRAs have distinct ranks to flexibly accommodate different M3IC tasks and inputs. These novel elements naturally necessitate a new initialization strategy.

\subsection{Multimodal Fusion in a Broader Context} \label{ssec:related-broader}



The multimodal fusion literature commonly partitions existing methods into three categories according to the level of integration: early fusion at the input level, intermediate fusion at the feature level, and late fusion at the decision level \cite{zhao2024deep}. 

Early fusion typically requires multimodal inputs to be spatially aligned, \eg between  MRI views \cite{yu2025superlightnet} or between RGB and depth images \cite{yin2024dformer}. However, this requirement does not hold for many M3IC tasks. In retinal disease recognition, for instance, CFP and OCT are in fact spatially orthogonal.
By tokenizing and concatenating image and text inputs into a unified sequence, LLM-style early fusion leverages deep network architectures and vast amounts of paired training samples to achieve cross-modal alignment  \cite{team2024chameleon,lei2025scalability}. However, given the low-resource nature of the medical domain, such a data-intensive approach is ill-suited for M3IC.

For intermediate fusion, vector concatenation is a common approach, as adopted by DynMM \cite{DynMM} for movie genre classification and STiL \cite{stil} for tabular-image classification. SFusion, by contrast,  aggregates embeddings from different modalities via weighted summation, where the modality-specific weights are computed through multiple self-attention layers \cite{sfusion}. In both approaches, fusion is carried out after UIE.

Recent methods for late fusion aim to dynamically combine unimodal predictions from individual modalities based on their classification ability \cite{mmef,aug}. MMEF \cite{mmef} estimates this ability via learned reliability coefficients, where each unimodal prediction is modeled as a Dempster-Shafer mass function. AUG \cite{aug} develops a boosting-style training scheme to iteratively strengthen the classification ability of weaker modalities. By definition, late fusion falls under the delayed fusion paradigm discussed in \cref{sec:intro}.

\section{Method} \label{sec:method}
\begin{figure*}[ht!]
    \centerline{
    \includegraphics[width=1.0\textwidth]{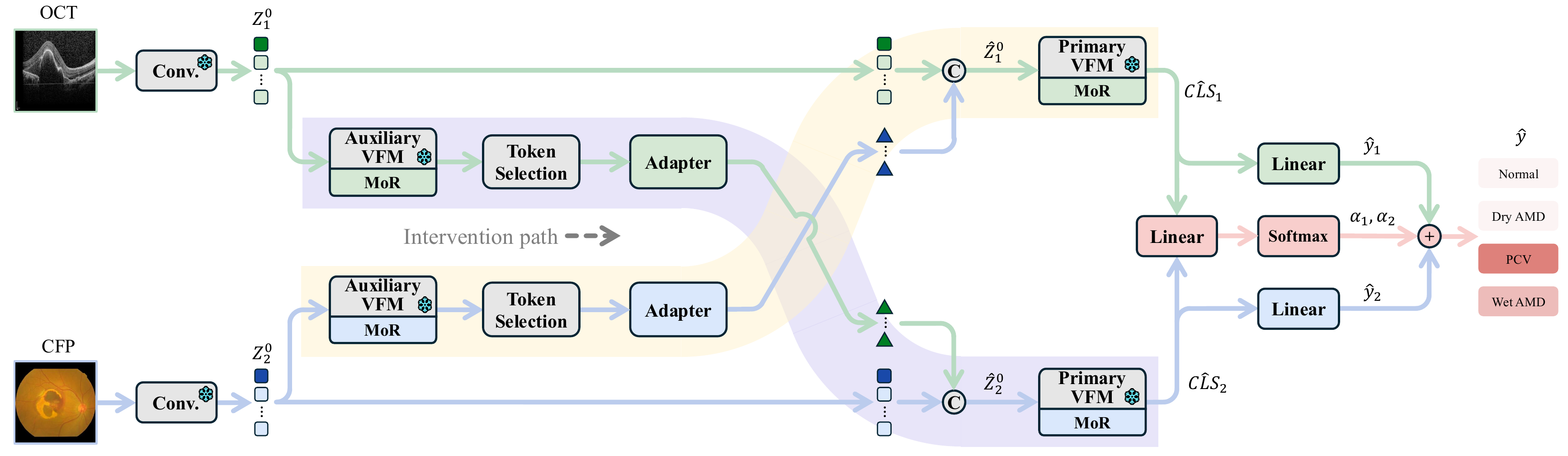}
    }
    \caption{\textbf{Proposed Early Intervention (\ourmodel) framework for M3IC}.
    Given a multimodal image sample (an OCT and a CFP in the showcase), each modality is designated in sequence as a \emph{target} modality, with the rest as its \emph{reference}. 
    \ourmodel utilizes the high-level semantics encapsulated in a few selected tokens from the reference as \textbf{intervention} (\intok) tokens to guide the target-modality feature extraction at an \emph{early} stage.
    Such a design mirrors a clinician's typical workflow, where an initial interpretation of one modality contextually informs the examination of another. Moreover, we develop \underline{M}ixture \underline{o}f varied-\underline{r}ank LoRAs (\textbf{MoR}) for parameter-efficient adaptation of VFMs to medical images. Our solution works for multiple M3IC tasks including retinal disease recognition, skin lesion recognition and knee anomaly classification. 
    }
    \label{fig:architecture}
\end{figure*}

We aim for a VFM-based medical image classifier that performs $C$-class disease or lesion recognition based on $M$-modality image input. Let $\mathcal{D}$ be a labeled set of $M$-modality  samples \{$(\mathbf{x}, y)$\}, where $\mathbf{x}[i]$ denotes the $i$-th modality image, $i=1, \ldots, M$, and $y \in \{0,1\}^C$ is the corresponding label in a $C$-class setting. In the proposed  \ourmodel framework, each modality is designated in sequence as a \emph{target} modality, indexed by $t$, with the rest as its \emph{reference} denoted as $R=\{1,\ldots,M\}\setminus\{t\}$. Thus, all the modalities will alternate between the role of target and reference. 

Given $\phi$ as a pre-trained VFM, we associate each modality with a primary VFM $\phi_p$ and an auxiliary VFM $\phi_a$. The function of the VFMs depends on the role of the current modality. When the modality acts as the target, $\phi_p$ extracts features for classification. When the modality serves as (part of) the reference, $\phi_a$ generates intervention (\intok) tokens. Next, we depict the generation of the \intok tokens (\cref{ssec:generate-int-token}), followed by their use in target-modality feature extraction (\cref{ssec:use-int-token}).  The proposed MoR module for parameter-efficient VFM adaptation is presented in \cref{ssec:mor}. Given the features extracted from the $M$ modalities, a lightweight late fusion module is provided in \cref{ssec:fusion}. A conceptual diagram of the \ourmodel framework is shown in \cref{fig:architecture}, with its training algorithm given in \cref{ssec:training}.

\subsection{\intok Token Generation} \label{ssec:generate-int-token}

For each reference modality, indexed by $r\in R$, we use an auxiliary VFM $\phi_{a,r}$ to generate its \intok tokens. Current VFMs, \eg CLIP \cite{clip} and DINOv2 \cite{dinov2}, are Vision Transformers (ViT) based. A ViT with $L$ transformer layers starts with a convolutional layer (Conv) that encodes an input image into a sequence of patch-level tokens $Z^0$. Prepended with a special CLS token, $Z^0$ is then fed into the stacked transformers, layer by layer, to progressively yield higher-level semantic tokens. For the ease of reference, we denote the output of the $l$-th transformer layer as $Z^l = \phi(Z^0, l)$, with its CLS token and patch tokens\footnote{To simplify our notation, we omit position embeddings and DINOv2's register tokens.} accessed via $Z^l[0]$ and $Z^l[1:]$, respectively.

Under the hypothesis that the output of the last transformer layer, \ie $Z^L$, best represents the reference modality's semantics, we opt to construct the \intok tokens from $Z^L$.
In particular, the \intok tokens consist of the CLS token, \ie $Z^L[0]$, typically known as a holistic representation, and a small number of $p$ contextual tokens derived from the patch tokens  $Z^L[1:]$. 

The contextual tokens are obtained by a single round of K-means clustering on the patch tokens, as outlined below. First, $p$ initial cluster centers are selected in a greedy manner based on the K-means++ initialization \cite{kmeans_pp}. Specifically, the first center is set to be the mean of all patch tokens. Subsequently, each new center is chosen as the patch token that maximizes the distance to the already selected centers. This greedy procedure ensures that the $p$ centers are well distributed across the feature space. The remaining patch tokens are assigned to their closest centers. The mean of each resultant cluster is then used as a differentiable contextual token. We set $p=4$, yielding negligible computational overhead.

As the \intok tokens are designed to intervene in the target modality's feature extraction at an early stage, 
an adapter is used to enhance the inter-modality compatibility.
More formally, the \intok sequence is generated as
\begin{equation} \label{eq:gen-int-tok}
\left\{
\begin{array}{ll}
Z_r^L & \leftarrow \phi_{a,r}(\mbox{Conv}(\mathbf{x}[r]), L), \\
\mbox{INT}_r & \leftarrow \mbox{Adapter}(Z^L_r[0]; \underbrace{\mbox{Select}(Z^L_r[1:], p)}_{\mbox{one-round K-means}}), \\
\mbox{INT} & \leftarrow \{\mbox{INT}_r | r\in R \},\\
\end{array}
\right.
\end{equation}
where $\mbox{INT}_r$ indicates the $1{+}p$ \intok tokens from the reference modality $r$. We use a two-layer MLP as the adapter.


\subsection{Primary Feature Extraction with \intok Tokens} \label{ssec:use-int-token}

For the current target modality ($t$), we use a primary VFM $\phi_{p,t}$ to extract features for classification.
In order to leverage the \intok token sequence at the very beginning of primary feature extraction, we concatenate the sequence with the output of the conv layer, namely $Z_t^0$, forming $\hat{Z}_t^0$. Using $\hat{Z}_t^0$ instead of $Z_t^0$ for subsequent forward computation in $\phi_{p,t}$ allows the \intok tokens to interact with and thus influence the target-modality tokens at all layers, ultimately yielding an \intok-intervened CLS token $\hat{\mbox{CLS}}_t$. Primary feature extraction with the \intok tokens is expressed as
\begin{equation} \label{eq:use-int-tok}
\left\{
\begin{array}{ll}
\hat{Z}_t^0 & \leftarrow \mbox{Concat}( \mbox{Conv}(\mathbf{x}[t]), \mbox{INT}), \\
\hat{\mbox{CLS}}_t & \leftarrow \phi_{p,t}(\hat{Z}_t^0, L)[0].
\end{array}
\right.
\end{equation}

Executing \cref{eq:use-int-tok} for each modality yields a set of multimodal features \{$\hat{\mbox{CLS}}_1, \ldots, \hat{\mbox{CLS}}_M$\}. These features will be combined for classification (\cref{ssec:fusion}).

\subsection{MoR for Parameter-Efficient VFM Adaptation} \label{ssec:mor}



\begin{figure}[!htbp]
  \centering
  \subfloat[Conference version \cite{EI}]{
    \includegraphics[width=0.9\columnwidth]{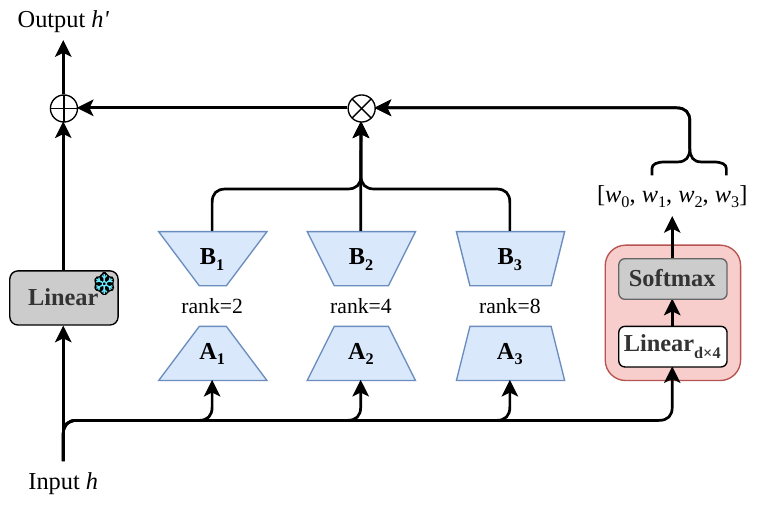}
    \label{fig:mor-a}
  }
  \vfill
  \subfloat[Journal version]{
    \includegraphics[width=\columnwidth]{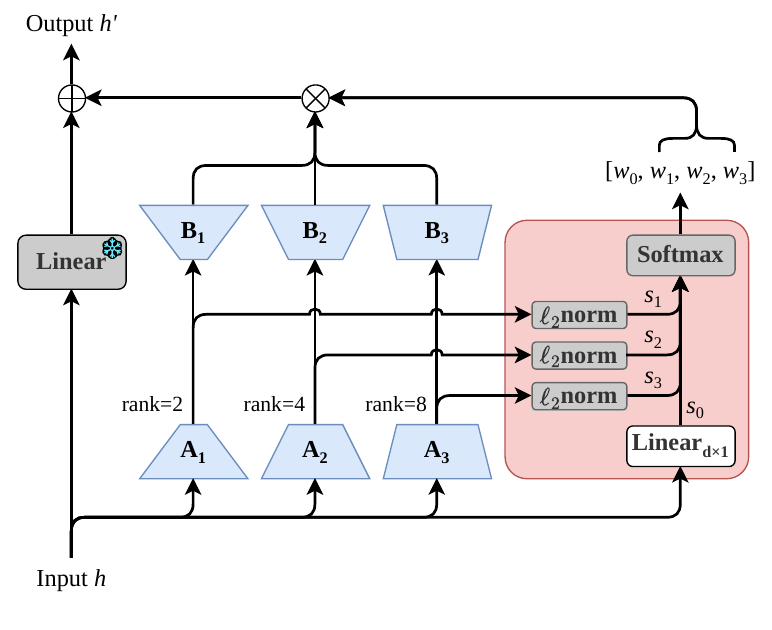}
    \label{fig:mor-b}
  }
  \caption{\textbf{Proposed MoR module}. It employs multiple LoRAs with different ranks instead of a fixed-value rank, and provide a \emph{bypass} path ($w_0$) to (partially) accept or reject the adaptation per instance. Compared to our conference version that uses a standard router with $4 \times d$ parameters, the journal extension is more parameter-efficient  with only $d$ parameters. 
  }
  \label{fig:mor}
\end{figure}

As shown in \cref{fig:mor}, our MoR module has three novel designs. First, \textbf{mixture of varied-rank LoRAs}. Based on our observation that the optimal choice of the rank used in LoRA varies across tasks and inputs, we leverage a mixture of LoRA experts, each with a distinct rank, instead of a fixed-value rank. In this work, we use three experts with ranks of 2, 4 and 8, respectively. These experts are then combined in an input-dependent manner, either via a standard router (\ie a \texttt{linear} layer followed by \texttt{softmax}, see \cref{fig:mor-a}) or through the parameter-efficient alternative described below (\cref{fig:mor-b}).


Second, \textbf{a parameter-efficient router with a \emph{bypass}}. A standard MoE router constrains its output weights to sum to unity \cite{moe}, enforcing an adaptation to a layer even when the original weights are better. We relax this constraint with a \emph{bypass} pathway that allows the layer to (partially) accept or reject the output of the individual LoRAs. In the extreme case where the bypass weight ($w_0$ in \cref{fig:mor}) is projected to be 1, the adaptation will be entirely bypassed. 

Following the typical use of LoRA, MoR  is also applied to the linear layers. 
As indicated above, MoR consists of three LoRAs. Each LORA, assigned with  a distinct rank $r_k \in \{2, 4, 8\}$, is parameterized by two low-rank factors $A_k \in \mathbb{R}^{r_k \times d}$ and $B_k \in \mathbb{R}^{d \times r_k}$. The adapted version of $h'$ is obtained by combining the output of the three LoRAs: 
\begin{equation} \label{eq:mor}
\left\{
\begin{array}{ll}
h_k & \leftarrow A_k h, \quad k = 1, 2, 3,\\
{}[w_0, w_1, w_2, w_3] & \leftarrow \mbox{PE-Router}(h, h_1, h_2, h_3),\\
h' & \leftarrow W h + \sum_{k=1}^3 w_k B_k h_k,\\
\end{array}
\right.
\end{equation}
where PE-Router indicates our parameter-efficient router. Since $\sum_{k=0}^3 w_k=1$, the bypass weight $w_0$ explicitly reduces the total weight allocated to the experts, enabling per-instance control over the adaptation strength.

Note that a standard router (\cref{fig:mor-a}) has $4 d$ parameters. Inspired by FlyLoRA \cite{flylora}, which uses the absolute value of each element in $Ah$ to measure the importance of the row vectors in $A$, we propose to use the $\ell_2$ norm of $A_kh$ to measure the importance of $A_k$ \wrt the input. Accordingly, the routing score of the $k$-th expert, denoted by $s_k$, can be computed \emph{without} parameters as $\|A_k h\|_2 / \sqrt{r_k}$, where $\sqrt{r_k}$ serves as a normalization scalar. Next, let $s_0$ be the routing score of the bypass path, which we obtain via a linear projection $g(h)$. The  weights are then obtained by applying softmax to $\{s_k\}$, see \cref{fig:mor-b}. PE-Router, with $d$ parameters only, is computed as 
\begin{equation} \label{eq:router}
\left\{
\begin{array}{ll}
s_0 & \leftarrow g(h), \\
s_k & \leftarrow \|h_k\|_2 / \sqrt{r_k},\quad k = 1, 2, 3,\\
{}[w_0, w_1, w_2, w_3] & \leftarrow \mbox{softmax}([s_0,\, s_1,\, s_2,\, s_3]).\\
\end{array}
\right.
\end{equation}

Third, 
\textbf{\underline{n}ear-\underline{z}ero start with \underline{n}on\underline{z}ero low-rank factors (NZ2)}.  Per LoRA expert, we aim for a \emph{nonzero} initialization of its factors $A$ and $B$ (the expert subscript $k$ is omitted for brevity) that meet two demands. First, a \emph{near}-zero start, \ie $B \times A \approx \mathbf{0}$, letting PEFT begin with minimal perturbation to the model. Second, a near full-row-rank $A$ that benefits both LoRA training and the $A$-norm based routing. Our proposed initialization, dubbed NZ2, runs in two steps. Firstly, we construct a sparse matrix $\hat{A} \in \mathbb{R}^{r \times d}$: the first $r/2$ rows of $\hat{A}$ are drawn from $\mathcal{N}(0, \sigma^2)$ with $\sigma=0.05$, and the remaining rows sampled from $\mathcal{N}(0, \epsilon^2)$ with $\epsilon = 1\mathrm{e}^{-5}$. Another sparse matrix $\hat{B}$ is constructed in a column-major fashion, with the first $r/2$ columns from $\mathcal{N}(0, \epsilon^2)$  and the remaining columns from $\mathcal{N}(0, \sigma^2)$, such that $\hat{B}\hat{A} \approx \mathbf{0}$. Next, to increase the rank of $\hat{A}$ while preserving its near-orthogonality to $\hat{B}$, we obtain an orthogonal matrix $Q$ via QR decomposition on a $r\times r$ random matrix, and set $A=Q\hat{A}$ and $B=\hat{B}Q^T$. This yields nonzero $A$ and $B$, which are nearly orthogonal and full-rank.

\subsection{Simple Late Fusion} \label{ssec:fusion}

With \ourmodel-enhanced UIE, multimodal feature fusion can be done in a simple manner (\cref{fig:architecture}).
Each modality-specific feature $\hat{\mbox{CLS}}_t$ is projected into a $C$-dimensional vector $\hat{y}_t$ by a linear layer.  A gating layer (linear followed by softmax) generates an $M$-dimensional weight vector $\{\alpha_1,\ldots,\alpha_M\}$,  representing the importance of each modality. The final prediction $\hat{y}$ is then computed as a weighted combination of the modality-specific predictions, see \cref{eq:fusion}.
\begin{equation} \label{eq:fusion}
\left\{
\begin{array}{ll}
\{\hat{y}_1, \ldots, \hat{y}_M\} & \leftarrow \{\mbox{Linear}(\hat{\mbox{CLS}}_t)|t=1,\ldots,M\},  \\
\{\alpha_1,\ldots,\alpha_M\} & \leftarrow \mbox{Gating}(\{\hat{\mbox{CLS}}_t\}),
\\
\hat{y} & \leftarrow \sum_{t=1}^M \alpha_t \hat{y}_t. 
\end{array}
\right.
\end{equation}
We use $\hat{y}$ as the classification output. 

\subsection{Training Algorithm} \label{ssec:training}

\textbf{Primary classification loss}. Given the modality-specific predictions $\{\hat{y}_t\}$ and the multimodal prediction $\hat{y}$ calculated by \cref{eq:fusion}, we define our primary loss $\mathcal{L}_p$ as the sum of the standard cross-entropy (CE) loss per prediction:
\begin{equation} \label{eq:primary-loss}
\mathcal{L}_p = \mbox{CE}(y, \hat{y}) +  \sum_{t=1}^M \mbox{CE}(y, \hat{y}_t).
\end{equation}

\textbf{Auxiliary classification loss}. 
In order to supervise the auxiliary VFMs $\{\phi_{a,1}, \ldots, \phi_{a,M}\}$ to generate task-related \intok tokens, we add a linear classification head to $\phi_a$. Denoting the output as $\hat{y}_a$, we define an auxiliary classification loss \wrt the auxiliary VFMs $\mathcal{L}_{ac}$ as
\begin{equation} \label{eq:aux-vfm-loss}
\left\{
\begin{array}{ll}
\mbox{CLS}_{a,i} & \leftarrow \phi_{a,i}(\mbox{Conv}(\mathbf{x}[i]), L)[0], \\
\hat{y}_{a,i} & \leftarrow \mbox{Linear}(\mbox{CLS}_{a,i}), \\
\mathcal{L}_{ac} & \leftarrow \sum_{i=1}^M \mbox{CE}(y, \hat{y}_{a,i}). 
\end{array}
\right.
\end{equation}

\textbf{Auxiliary gating loss}. 
Ideally, the gating module as defined in \cref{eq:fusion} should learn to produce optimal weights for late fusion. However, we empirically observe that the VFM-based framework tends to overfit the training data, causing the modality-specific VFMs to saturate and perform nearly equally well in the training stage. This saturation obscures their true performance differences, making it difficult for the gating network to discern which VFM is genuinely better. To alleviate the issue, we introduce a modality-wise prior $\pi$, derived with ease from the best-performing modality on the training set, to supervise the gating layer. Specifically, $\pi$ is a one-hot vector indicating the winning modality. We thus define an auxiliary gating loss $\mathcal{L}_{ag}$ as:
\begin{equation}\label{eq:aux-gating-loss}
\mathcal{L}_{ag} = \mbox{CE}(\pi, \{\alpha_1, \ldots, \alpha_M\}).
\end{equation}

The whole network is end-to-end trained by minimizing a joint loss as
\begin{equation} \label{eq:total-loss}
\mathcal{L}_{p} + \lambda_1 \mathcal{L}_{ac} + \lambda_2 \mathcal{L}_{ag},
\end{equation}
where $\lambda_1$ and $\lambda_2$ are hyper-parameters, with default values of 0.3 and 0.1, respectively. 
See \cref{alg:training} for the training procedure.

\begin{algorithm}[!htbp]
\caption{\ourmodel in a PyTorch style}
\label{alg:training}
\begin{PythonB}[mathescape=true]
Input: Multimodal training_set D={(x,y)},
       VFM phi, Modality prior pi
Output: Trained EI network

L, p = 12, 4
MoR_a, MoR_p = [MoR]*M, [MoR]*M
phi.requires_grad = False
EI = [MoR_a, MoR_p, adapter, linear, ...]
optimizer = SGD(EI.parameters)
# Show only one training epoch
for mini-batch {(x, y)} in D:
    optimizer.zero_grad()
    # Extract auxiliary CLS and patch tokens
    cls_a, pat_a = [0]*M, [0]*M
    for r in range(M):
        phi_a = phi, MoR_a[r]
        Z_L = phi_a(conv(x[r]), L)
        cls_a[r] = Z_L[0]
        pat_a[r] = K_means(Z_L[1:], p)
    # Early intervention
    cls_p = [0]*M
    for t in range(M):
        INT = []
        R = [r for r in range(M) if r!=t]
        for r in R:
            INT += adapter(cls_a[r], pat_a[r])
        Z_0 = concat(conv(x[t]), INT)
        phi_p = phi, MoR_p[t]
        cls_p[t] = phi_p(Z_0, L)[0]
    # Simple late fusion
    alpha = gating(cls_p)
    hat_yp = linear(cls_p)
    hat_y = torch.dot(alpha, hat_yp)
    # Loss computation
    loss_p = ce(y, hat_y) + ce(y, hat_yp)
    loss_ac = ce(y, linear(cls_a))
    loss_ag = ce(pi, alpha)
    loss = loss_p+0.3*loss_ac+0.1*loss_ag
    loss.backward()
    optimizer.step()
\end{PythonB}
\end{algorithm}

\section{Experiments} \label{sec:eval}

\subsection{Experimental Setup}
\textbf{Datasets}.
We adopt three public medical image datasets corresponding to three distinct organs: \\
$\bullet$ MMC-AMD \cite{jbhi22-mmcamd} : Four-class age-related macular degeneration (AMD) classification, \ie \emph{dry AMD (dAMD)}, \emph{wet AMD (wAMD)}, \emph{polypoidal choroidal vasculopathy (PCV)} and \emph{normal}. Each multimodal sample consists of a CFP image and an OCT image. \\
$\bullet$ Derm7pt \cite{derm7pt} : Five-class skin lesion recognition, \ie \emph{basal cell carcinoma} (\emph{BCC}), \emph{nevus} (\emph{NEV}), \emph{melanoma} (\emph{MEL}), \emph{seborrheic keratosis} (\emph{SK}) and \emph{miscellaneous} (\emph{MISC}). Each sample consists of a dermoscopic image and a clinic image. \\
$\bullet$ MRNet \cite{mrnet} : Multi-label knee abnormality classification, \ie \emph{abnormal}, \emph{anterior cruciate ligament tear (ACLT)} and \emph{meniscal tear (MT)}. Each multimodal sample consists of MRI images from sagittal-view, axial-view and coronal-view.

For MMC-AMD and Derm7pt, we use their official data splits.  
As for MRNet, since 
only the training and validation sets are publicly available, we use the validation set as our test set, whilst selecting at random 120 samples from the training set as a held-out validation set. An overview of the datasets is given in \cref{tab:dataset}. See visual examples in \cref{fig:att}.

\begin{table}[!htbp]
\caption{\textbf{Three multimodal medical image datasets used in our experiments}, with $M$ and $C$ indicate the number of modalities and the number of classes, respectively.}
\label{tab:dataset}
\centering
\renewcommand{\arraystretch}{1.2}
\resizebox{0.95\linewidth}{!}{
\begin{tabular}{@{}llrrrrrr@{}}
\toprule
\multirow{2}{*}{\textbf{Dataset}} & \multirow{2}{*}{\textbf{Organ}} & \multirow{2}{*}{$M$} & \multirow{2}{*}{$C$} & \multicolumn{4}{c}{\#\textbf{Multimodal samples}}\\
\cline{5-8}
&&&& \emph{Total} & \emph{Train} & \emph{Val.} & \emph{Test} \\
\hline
MMC-AMD\cite{jbhi22-mmcamd} & eye & 2 & 4 & 768 & 610 & 79 & 79 \\
Derm7pt\cite{derm7pt}  & skin & 2 & 5 &1,011 & 413 & 203 & 395 \\
MRNet\cite{mrnet}  & knee & 3 & 3 & 1,250 & 1,010 & 120 & 120 \\
\bottomrule
\end{tabular}
}
\end{table}

\textbf{Evaluation criteria}. We adopt Average Precision (AP) as the primary performance metric, with AUC and the harmonic mean of sensitivity and specificity (S2) as secondary metrics. Each metric is computed per class and averaged over classes, denoted as mAP, mAUC, and mS2, respectively. Unless otherwise stated, we report mAP, with the other metrics provided in the supplementary material. 

\textbf{Choice of VFMs}. We adopt two generic VFMs: CLIP(ViT-B/16) \cite{clip} and  
DINOv2(ViT-B/14) \cite{dinov2}.  See \cref{sssec:eval-vfm} for a comparison to domain-specific counterparts.



\textbf{Details of implementation}. The input image resolutions are set according to the official recommendations for each dataset: 448$\times$448 for MMC-AMD, 512$\times$512 for Derm7pt, and 256$\times$256 for MRNet.
Per 3D-volume sample in MRNet, we select four slices, which are sampled at random during training and sampled uniformly for inference, respectively. A sample-level feature is obtained by mean pooling over selected slices.

We use the following training protocol.
The network optimizer is SGD with a momentum of 0.95 and a weight decay of 1e-4. 
The learning rate scheduler is CyclicLR (cycling between 1e-5 and 1e-3) combined with a warmup strategy.
Batch size is set to 8. 
Validation is performed every epoch. Early stop occurs if there is no performance increase in 10 successive validations. Unless otherwise specified, for each modality we initialize its auxiliary and primary VFMs using a model pre-adapted with MoR on the unimodal training data. All reported results are the mean of three training runs.
All experiments were conducted using PyTorch 2.3 on four NVIDIA GeForce RTX 4090 GPUs and four 5880 GPUs.

\subsection{\ourmodel \emph{versus} Previous Methods}  \label{ssec:eval-sota}

\begin{table*}[!htbp]
\caption{\textbf{M3IC Performance of different methods across three datasets}. Methods are sorted in ascending order by their MEAN performance. The amount of trainable parameters per method is averaged over the datasets. \emph{N}: \emph{Normal}. \emph{A}: \emph{Abnormal}.}
\label{tab:sota}
\renewcommand{\arraystretch}{1.2}
\resizebox{\linewidth}{!}{
\begin{tabular}{@{}lrrrrrrrrrrrrrrrrrrr@{}}
\toprule
\multirow{2}{*}{\textbf{Method}} & \multirow{2}{*}{\textbf{\makecell{Params.\\(M)}}} & \multirow{2}{*}{\textbf{MEAN}} & \multicolumn{5}{c}{\textbf{MMC-AMD}}&& \multicolumn{6}{c}{\textbf{Derm7pt}}&& \multicolumn{4}{c}{\textbf{MRNet}} \\
\cline{4-8} \cline{10-15} \cline{17-20}

&&& mAP & \emph{N} & \emph{dAMD} & \emph{PCV} & \emph{wAMD} && mAP & \emph{BCC} & \emph{NEV} & \emph{MEL} & \emph{SK} & \emph{MISC} && mAP & \emph{A} & \emph{MT} & \emph{ACLT} \\

\hline

\multicolumn{2}{@{}l}{VFM: \textbf{CLIP}} \\

STiL {\scriptsize CVPR25}\cite{stil} &216.2& 0.640 & 0.787 & 1.0 & 0.925 & 0.718 & 0.504 && 0.362 & 0.154 & 0.777 & 0.477 & 0.076 & 0.325 && 0.773 & 0.921 & 0.668 & 0.729\\
AUG {\scriptsize NeurIPS25}\cite{aug} &201.4& 0.641 & 0.707 & 1.0 & 0.706 & 0.476 & 0.645 && 0.385 & 0.278 & 0.805 & 0.406 & 0.087 & 0.349 && 0.831 & 0.957 & 0.701 & 0.834 \\

RadDiag {\scriptsize NC24}\cite{RadDiag} & 212.2 & 0.648 & 0.783 & 1.0 & 0.922 & 0.659 & 0.551&& 0.352 & 0.128 & 0.756 & 0.480 & 0.068 & 0.327&& 0.809 & 0.954 & 0.652 & 0.822 \\
MMEF {\scriptsize InfFus25}\cite{mmef} &201.0& 0.661 & 0.828 & 1.0 & 0.897 & 0.755 & 0.659 && 0.412 & 0.196 & 0.824 & 0.540 & 0.141 & 0.361 && 0.743 & 0.946 & 0.634 & 0.650 \\

DynMM {\scriptsize CVPRW23}\cite{DynMM} & 402.2 & 0.671 & 0.830 & 1.0 & 0.928 & 0.810 & 0.580&& 0.349 & 0.148 & 0.767 & 0.422 & 0.063 & 0.345&& 0.833 & 0.958 & 0.696 & 0.846 \\

MM-MIL {\scriptsize MM21}\cite{mm-mil} & 202.5 & 0.671 & 0.818 & 1.0 & 0.919 & 0.761 & 0.592&& 0.360 & 0.105 & 0.782 & 0.441 & 0.075 & 0.396&& 0.835 & 0.972 & 0.674 & 0.858 \\

CosCat {\scriptsize ArtMed25}\cite{CosCatNet} & 201.0 & 0.679 & 0.839 & 1.0 & \underline{0.929} & 0.826 & 0.602&& 0.364 & 0.163 & 0.775 & 0.493 & 0.080 & 0.309&& 0.835 & 0.974 & 0.685 & 0.845 \\

SFusion {\scriptsize MICCAI23}\cite{sfusion} & 263.4 & 0.683 & 0.840 & 1.0 & \textbf{0.946} & 0.754 & 0.662&& 0.375 & 0.158 & 0.737 & 0.523 & 0.067 & 0.391&& 0.835 & 0.976 & 0.673 & 0.856 \\

FPT+ {\scriptsize TNNLS25}\cite{fptp} & 2.1  & 0.693 & 0.736 & 1.0 & 0.873 & 0.647 & 0.424 &  & 0.500 & 0.329 & 0.856 & 0.533 & 0.267 & 0.513 &  & 0.843 & 0.973 & 0.727 & 0.828\\

LateLoRA {\scriptsize MIDL25}\cite{latelora}& 0.2  & 0.718 & 0.815 & 1.0 & 0.875 & 0.802 & 0.582 &  & 0.508 & 0.342 & 0.875 & 0.477 & 0.273 & 0.572 &  & 0.831 & 0.973 & 0.738 & 0.782 \\

MMRAD {\scriptsize InfFus25}\cite{mmrad} & 9.3 & 0.729 & 0.819 & 1.0 & 0.928 & 0.684 & 0.664&& 0.549 & 0.421 & 0.879 & 0.510 & 0.345 & 0.590&& 0.818 & 0.959 & 0.679 & 0.815 \\

LKA {\scriptsize MICCAI25}\cite{lka} & 1.4  & 0.752 & 0.846 & 1.0 & 0.923 & 0.841 & 0.619 &  & 0.574 & 0.375 & 0.897 & 0.639 & 0.338 & 0.621 &  & 0.836 & \underline{0.978} & 0.727 & 0.804 \\

FreqFiT {\scriptsize MICCAI25}\cite{freqfit} & 18.1 & 0.768 & 0.865 & 1.0 & 0.928 & 0.870 & 0.660 &  & 0.602 & 0.463 & 0.918 & 0.613 & 0.411 & 0.603 &  & 0.838 & 0.972 & 0.732 & 0.811 \\

MMRAD-MoR & 12.2 & 0.788 & 0.859 & 1.0 & 0.832 & 0.862 & 0.741 && 0.666 & 0.502 & \textbf{0.931} & 0.719 & 0.487 & 0.692 && 0.839 & 0.970 & 0.686 & 0.861 \\

MM-MIL-MoR & 4.7 & 0.808 & \underline{0.898} & 1.0 & 0.910 & \textbf{0.937} & 0.744 && 0.679 & 0.512 & \underline{0.930} & \textbf{0.761} & 0.501 & 0.692 && 0.846 & \underline{0.978} & 0.699 & 0.862\\

\ourmodel (\emph{conf.}) & 8.9 & \underline{0.822} & 0.889 & 1.0 & 0.880 & 0.896 & \underline{0.782} && \underline{0.715} & \underline{0.547} & 0.928 & \underline{0.758} & \textbf{0.647} & \underline{0.694} && \underline{0.861} & 0.973 & \underline{0.748} & \underline{0.863} \\

\rowcolor{blue!20}
\ourmodel & 8.2 & \textbf{0.835} & \textbf{0.907} & 1.0 & 0.901 & \underline{0.932} & \textbf{0.796} && \textbf{0.729} & \textbf{0.679} & 0.925 & 0.756 & \underline{0.576} & \textbf{0.707} && \textbf{0.868} & \textbf{0.980} & \textbf{0.751} & \textbf{0.872}\\

\hline

\multicolumn{2}{@{}l}{VFM: \textbf{DINOv2}} \\
AUG  & 202.5 & 0.684 & 0.794 & 1.0 & 0.910 & 0.506 & 0.762 && 0.420 & 0.224 & 0.838 & 0.514 & 0.074 & 0.453 && 0.837 & 0.970 & 0.711 & 0.831 \\
SFusion  & 265.8 & 0.687 & 0.821 & 1.0 & 0.915 & 0.678 & 0.690&& 0.401 & 0.208 & 0.821 & 0.462 & 0.077 & 0.438&& 0.839 & 0.972 & 0.720 & 0.826 \\
DynMM  & 404.3 & 0.688 & 0.854 & 1.0 & 0.933 & 0.782 & 0.701&& 0.377 & 0.183 & 0.813 & 0.427 & 0.068 & 0.395&& 0.833 & 0.970 & 0.689 & 0.839 \\
MMEF  & 202.0 & 0.689 & 0.847 & 1.0 & 0.924 & 0.792 & 0.673 && 0.422 & 0.191 & 0.806 & 0.570 & 0.099 & 0.445 && 0.799 & 0.888 & 0.728 & 0.781 \\

STiL  & 217.2 & 0.690 & 0.782 & 1.0 & 0.883 & 0.652 & 0.593 && 0.446 & 0.334 & 0.858 & 0.433 & 0.112 & 0.492 && 0.842 & 0.974 & 0.666 & \textbf{0.887}\\
CosCat  & 202.0 & 0.693 & 0.823 & 1.0 & 0.931 & 0.767 & 0.594&& 0.417 & 0.267 & 0.836 & 0.447 & 0.081 & 0.451&& 0.841 & 0.973 & 0.693 & \underline{0.856} \\
RadDiag  & 213.3 & 0.717 & 0.826 & 1.0 & 0.935 & 0.708 & 0.662&& 0.499 & 0.365 & 0.847 & 0.554 & 0.182 & 0.549&& 0.825 & 0.962 & 0.691 & 0.822 \\

FPT+     & 2.1  & 0.731 &  0.773 & 1.0 & 0.899 & 0.761 & 0.430 &  & 0.576 & 0.519 & 0.883 & 0.562 & 0.385 & 0.534 &  & 0.844 & 0.975 & 0.749 & 0.807 \\

MM-MIL  & 203.6 & 0.733 & 0.863 & 1.0 & 0.932 & 0.861 & 0.658&& 0.502 & 0.406 & 0.867 & 0.561 & 0.170 & 0.507&& 0.835 & 0.972 & 0.707 & 0.826 \\
MMRAD  & 9.3 & 0.735 & 0.821 & 1.0 & 0.919 & 0.789 & 0.576&& 0.566 & 0.437 & 0.879 & 0.521 & 0.446 & 0.546&& 0.818 & 0.968 & 0.705 & 0.781 \\

LateLoRA & 0.2 & 0.758 &  0.821 & 1.0 & 0.862 & 0.777 & 0.645 &  & 0.634 & 0.551 & 0.905 & 0.600 & 0.571 & 0.542 &  & 0.820 & 0.960 & 0.733 & 0.768 \\

FreqFiT  & 23.2 & 0.765 &  0.866 & 1.0 & \textbf{0.948} & 0.865 & 0.651 &  & 0.602 & 0.483 & 0.903 & 0.577 & 0.485 & 0.564 &  & 0.827 & 0.975 & 0.728 & 0.779 \\

LKA      & 1.4 & 0.788 &  0.869 & 1.0 & \underline{0.943} & 0.860 & 0.674 &  & 0.654 & 0.616 & 0.906 & 0.602 & 0.582 & 0.562 &  & 0.841 & \textbf{0.983} & \underline{0.763} & 0.777 \\

MMRAD-MoR & 12.2 & 0.807 & 0.870 & 1.0 & 0.862 & 0.919 & 0.699 && 0.713 & 0.605 & 0.910 & 0.731 & 0.765 & 0.553 && 0.838 & 0.963 & 0.759 & 0.790\\

MM-MIL-MoR & 4.7 & 0.812 & 0.864 & 1.0 & 0.854 & 0.880 & 0.721 && 0.745 & 0.675 & \underline{0.935} & \textbf{0.775} & 0.764 & 0.576 && 0.827 & 0.973 & 0.746 & 0.762\\

\ourmodel (\emph{conf.})  & 8.9 & \underline{0.841} & \underline{0.909} & 1.0 & 0.916 & \underline{0.929} & \underline{0.789} && \underline{0.767} & \underline{0.690} & 0.932 & 0.755 & \textbf{0.777} & \textbf{0.682}&& \underline{0.848} & \underline{0.978} & 0.734 & 0.833 \\

\rowcolor{blue!20}
\ourmodel & 8.2 & \textbf{0.854} & \textbf{0.925} & 1.0 & 0.907 & \textbf{0.950} & \textbf{0.843} && \textbf{0.786} & \textbf{0.799} & \textbf{0.938} & \textbf{0.775} & \underline{0.766} & \underline{0.652} && \textbf{0.851} & 0.965 & \textbf{0.764} & 0.823 \\

\bottomrule

\end{tabular}}
\end{table*}

\textbf{Vanilla baselines}.
To ensure reproducibility, we opt for open-source methods. 
Moreover, since our evaluation involves medical image classification tasks from three domains (\cref{tab:dataset}), the chosen baselines need to be applicable with minimal adjustment. For a more comprehensive comparison, we also include unimodal PEFT methods, \ie FPT+ \cite{fptp}, LKA \cite{lka}, FreqFiT \cite{freqfit} and LateLoRA \cite{latelora}, extending them to M3IC by applying late average fusion to their per-modality adapted outputs. Accordingly, we compile a list of 13 baselines, put into four groups based on the fusion methods used: 
\begin{itemize}
    \item \emph{Concatenation}: CosCat\footnote{\url{https://github.com/Zuo-Lihan/CosCatNet-Adaptive_Fusion_Algorithm}}  {\scriptsize ArtMed25} \cite{CosCatNet}, DynMM\footnote{\url{https://github.com/zihuixue/DynMM}} {\scriptsize CVPRW23} \cite{DynMM} and STiL\footnote{\url{https://github.com/siyi-wind/STiL}} {\scriptsize CVPR25}  \cite{stil}.
    \item \emph{Weighted summation}: MM-MIL\footnote{\url{https://github.com/WeiQijie/MM-RDC}} {\scriptsize MM21} \cite{mm-mil}, SFusion\footnote{\url{https://github.com/scut-cszcl/SFusion}} {\scriptsize MICCAI23} \cite{sfusion}, MMEF\footnote{\url{https://github.com/iWeisskohl/Deep-evidential-fusion}} {\scriptsize InfFus25} \cite{mmef} and AUG\footnote{\url{https://github.com/njustkmg/NeurIPS25-AUG}} {\scriptsize NeurIPS25} \cite{aug}.
    \item \emph{Attention}: RadDiag\footnote{\url{https://github.com/qiaoyu-zheng/RP3D-Diag}} {\scriptsize NC24}\cite{RadDiag} and MMRAD\footnote{\url{https://github.com/Jingtao-Li-CVer/MMRAD}} {\scriptsize InfFus25}\cite{mmrad}.
    \item \emph{Late fusion}: FPT+\footnote{\url{https://github.com/YijinHuang/FPT}} {\scriptsize TNNLS25} \cite{fptp}, LKA\footnote{\url{https://github.com/misswayguy/LKA}} \cite{lka} {\scriptsize MICCAI25}, FreqFiT\footnote{\url{https://github.com/tsly123/FreqFiT_medical}} {\scriptsize MICCAI25} \cite{freqfit} and LateLoRA\footnote{\url{https://github.com/computational-cell-analytics/peft-sam}} {\scriptsize MIDL25} \cite{latelora}. 
\end{itemize}
Note that for a fair comparison, for all baselines we replace their original visual backbones with the VFMs. We follow their training strategies, having the backbones either frozen (as in MMRAD which uses prompt learning or FPT+, LKA, FreqFit and LateLoRA that are parameter-efficient) or full-parameter fine-tuned (as in the rest).
See \cref{tab:sota} for the number of trainable parameters per baseline.

\textbf{Enhanced baselines}.
To enable a more conclusive comparison, we enhance  MM-MIL and MMRAD by adapting their VFMs with our MoR. We refer to these enhanced versions as \emph{MM-MIL-MoR} and \emph{MMRAD-MoR}, respectively. 



\textbf{Results}.
\cref{tab:sota} shows the per-dataset, per-class and overall results of the different methods. Among the vanilla baselines, the top performer depends on the VFM used. FreqFiT achieves the best results with CLIP, while LKA excels with DINOv2. \ourmodel outperforms both methods by a large margin in light of the overall performance, \ie 0.768$\rightarrow$0.835 with CLIP as VFM and 0.788$\rightarrow$0.854 with DINOv2 as VFM. The improvement is consistently observed across datasets.


Among the three M3IC tasks, Derm7pt (skin lesion recognition) appears to be the most challenging, as all methods achieve relatively lower performance on this dataset.
Again, \ourmodel clearly outperforms the best baselines (FreqFiT and LKA) in this task. 

The superior performance of MM-MIL-MoR and MMRAD-MoR over their original counterparts shows that MoR is more effective than both full-parameter tuning (used in MM-MIL) and prompt learning (used in MMRAD). 
Since these enhanced baselines also employ MoR, the better performance of \ourmodel further validates the advantage of our proposed early intervention strategy.


\subsection{Ablation Studies} \label{ssec:eval-abl}


To understand the influence of the individual components, we conduct extensive ablation studies, covering the acquisition and usage of the \intok tokens (\cref{sssec:eval-int}), contextual tokens (\cref{sssec:eval-kmeans-k}),  auxiliary losses (\cref{sssec:eval-aux-losses}), alternative PEFT methods to MoR (\cref{sssec:eval-peft}), MoR initialization (\cref{sssec:eval-mor-init}), MoR routing (\cref{sssec:eval-router}), and lastly the use of domain-specific VFMs (\cref{sssec:eval-vfm}).

\subsubsection{Choices of \intok Tokens Acquisition and Usage} \label{sssec:eval-int}
\begin{figure}[!htbp]
  \centering
  \subfloat[Acquisition]{%
    \includegraphics[width=0.49\columnwidth]{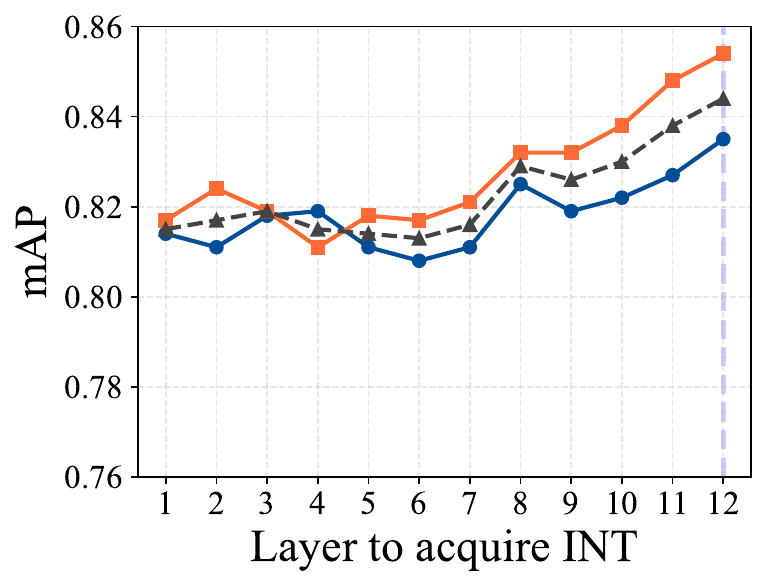}%
    \label{fig:int-acq}%
  }%
  \hfill
  \subfloat[Usage]{%
    \includegraphics[width=0.49\columnwidth]{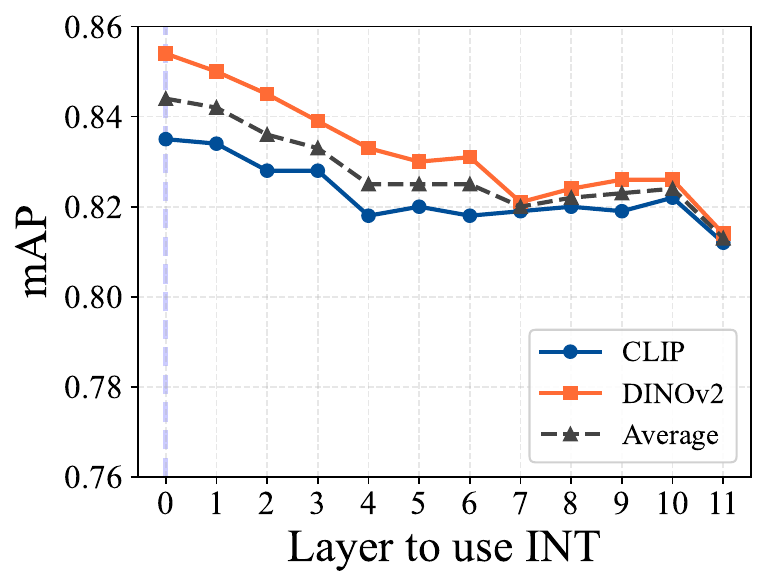}%
    \label{fig:int-use}%
  }
  \caption{\textbf{Performance curves \wrt \intok tokens acquisition and usage at different layers}. 
  }
  \label{fig:int}
\end{figure}

By default, we acquire the \intok tokens from the last layer of the auxiliary VFMs (\cref{ssec:generate-int-token}), and insert them into the input of the first Transformer layer of the target modality's primary VFM (\cref{ssec:use-int-token}). 
As \cref{fig:int} shows,
%
the last layer of the auxiliary VFM is the best position for acquiring the \intok tokens. Meanwhile, inserting these tokens at Layer 0 of the primary VFM, which is the earliest possible point, yields the best performance. This result 
clearly justifies the early intervention principle.

\subsubsection{Effect of the Contextual Tokens} \label{sssec:eval-kmeans-k}

\begin{figure}[!htbp]
  \centering
  \subfloat[MMC-AMD]{
    \includegraphics[width=\columnwidth]{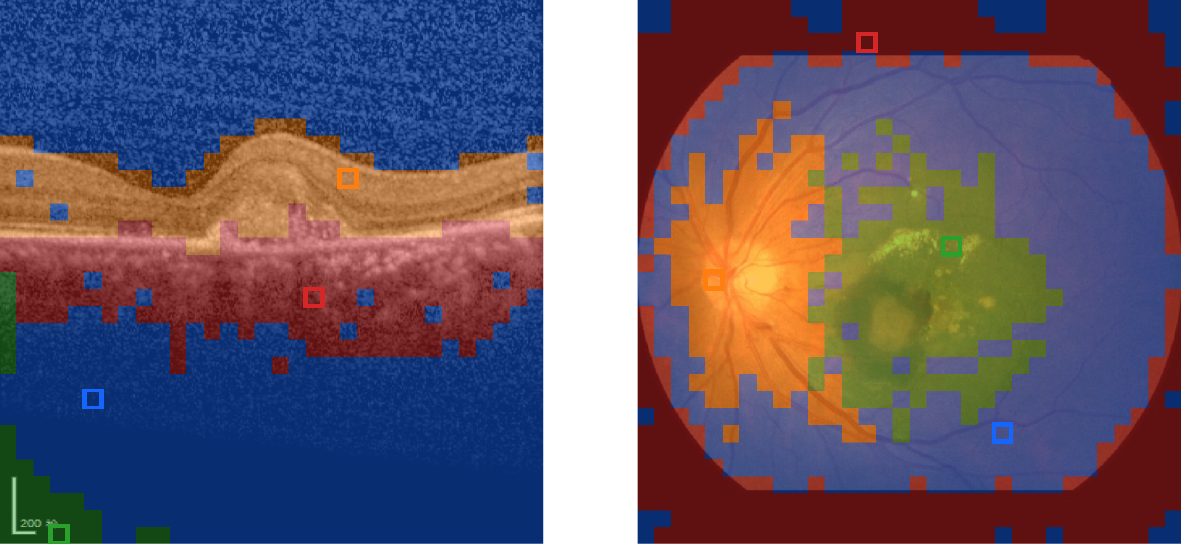}
    \label{fig:kmeans-mmc}
  }

  \subfloat[Derm7pt]{
    \includegraphics[width=\columnwidth]{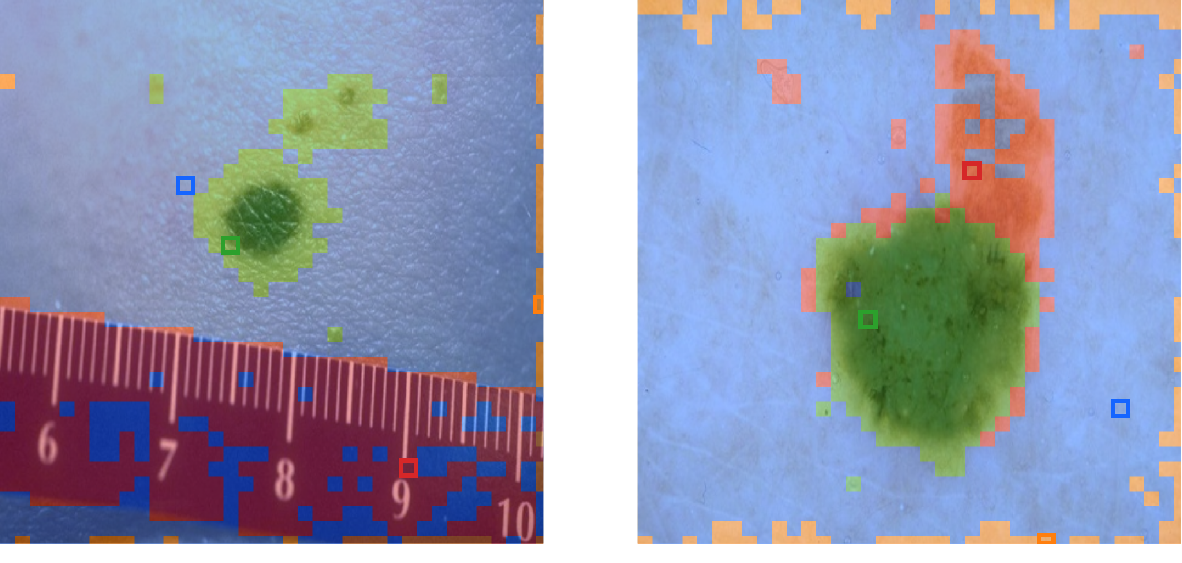}
    \label{fig:kmeans-derm}
  }

  \subfloat[MRNet]{
    \includegraphics[width=\columnwidth]{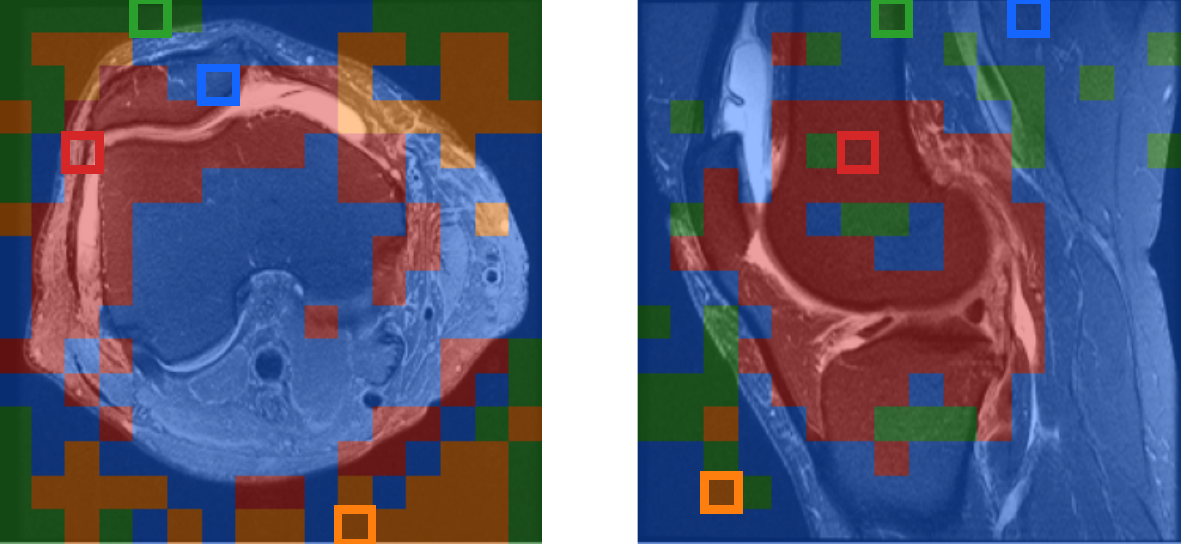}
    \label{fig:kmeans-mrnet}
  }
  \caption{
\textbf{Projecting contextual tokens onto image regions}. Given the four contextual tokens obtained, each patch is assigned to its closest contextual token and colored accordingly (red, green, blue, and yellow). Per token, the closest patch is further highlighted with a bold border. These highlighted patches span both normal and abnormal regions. Bested viewed in color.   
  }
  \label{fig:kmeans}
\end{figure}

As shown in \cref{tab:kmeans-k}, the inclusion of four contextual tokens yields a consistent, albeit modest, improvement over our conference version \cite{EI} that uses only the CLS token as the \intok token (\ie $p=0$). As visualized by \cref{fig:kmeans}, the contextual tokens provide a compact yet informative summary of the patch tokens from the reference modalities.


\begin{table}[!htbp]
\centering
\caption{\textbf{Effect of the number of contextual tokens} $p$. Setting $p=0$ means using only the CLS token as the \intok token. Relative improvement over this baseline is shown in parentheses.
}
\label{tab:kmeans-k}
\renewcommand{\arraystretch}{1.2}
\resizebox{0.75\linewidth}{!}
{
\begin{tabular}{@{}l lrrr@{}}
\toprule
\textbf{$p$} & \textbf{MEAN} & \emph{MMC-AMD} & \emph{Derm7pt} & \emph{MRNet} \\
\midrule
\multicolumn{2}{@{}l}{VFM: \textbf{CLIP}} \\
0 & 0.825 & 0.895 & \underline{0.720} & 0.862 \\
1 & 0.820  (-0.6\%$\downarrow$) & 0.896 & 0.709 & 0.856 \\
2 & 0.827  (+0.2\%$\uparrow$) & 0.898 & 0.715 & 0.867 \\
\rowcolor{blue!20}
4 & \textbf{0.835}  (+1.2\%$\uparrow$) & \underline{0.907} & \textbf{0.729} & \underline{0.868} \\
8  & \textbf{0.835}  (+1.2\%$\uparrow$) & \textbf{0.913} & 0.719 & \textbf{0.872}\\

\midrule
\multicolumn{2}{@{}l}{VFM: \textbf{DINOv2}} \\
0 & 0.847 & 0.909 & 0.781 & \textbf{0.851} \\
1 & 0.836 (-1.3\%$\downarrow$) & 0.906 & 0.768 & 0.835 \\
2 & 0.847 (0.0\%) & 0.915 & 0.784 & 0.841 \\
\rowcolor{blue!20}
4 & \textbf{0.854} (+0.8\%$\uparrow$) & \textbf{0.925} & \underline{0.786} & \textbf{0.851} \\
8 & \underline{0.853}  (+0.7\%$\uparrow$) & \underline{0.923} & \textbf{0.788} & 0.847 \\
\bottomrule

\end{tabular}}
\end{table}

\subsubsection{Necessity of the Auxiliary Losses} \label{sssec:eval-aux-losses}

\begin{figure}[!htbp]
  \centering
  \subfloat[Classification loss $\mathcal{L}_{ac}$]{%
    \includegraphics[width=0.49\columnwidth]{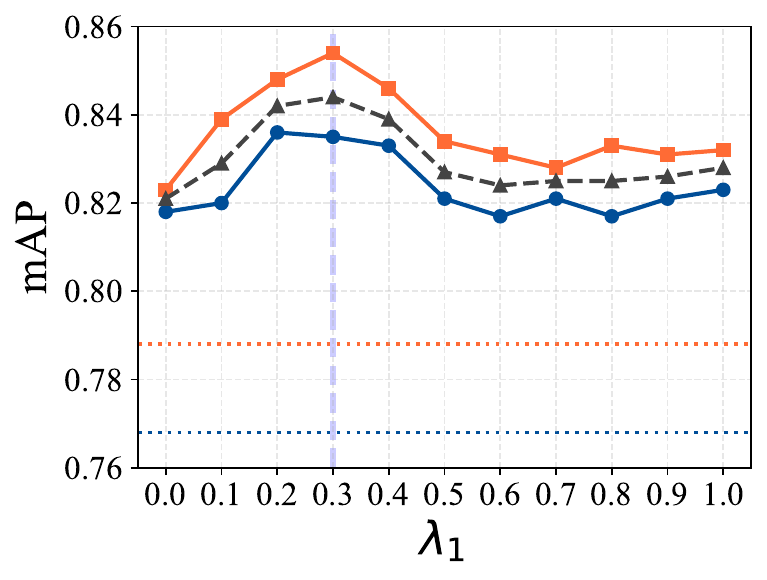}%
    \label{fig:lambda1}%
  }%
  \hfill
  \subfloat[Gating loss $\mathcal{L}_{ag}$]{%
    \includegraphics[width=0.49\columnwidth]{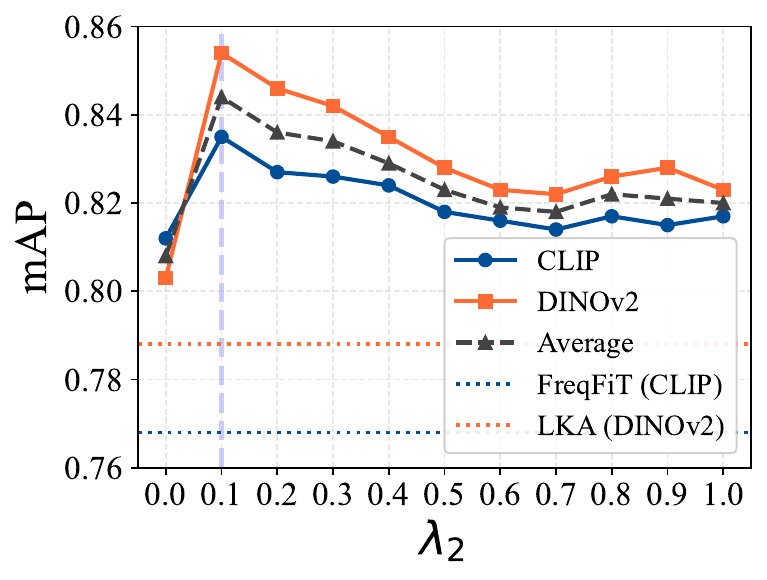}%
    \label{fig:lambda2}%
  }
  \caption{\textbf{Influence of the auxiliary losses}. 
  Disabling either loss by setting its weight to zero (the left ending point of the performance curve) clearly degrades the overall performance.}
  \label{fig:lambda}
\end{figure}

To investigate the impact of the two auxiliary losses $\mathcal{L}_{ac}$ and $\mathcal{L}_{ag}$, we conduct a sensitivity analysis by varying their weights, \ie  $\lambda_1$ for $\mathcal{L}_{ac}$ and $\lambda_2$ for $\mathcal{L}_{ag}$, one at a time. We set $\lambda_1 = 0.3$ when varying $\lambda_2$, and $\lambda_2 = 0.1$ when varying $\lambda_1$.
Disabling either loss (by setting its corresponding weight to zero) results in a clear degradation of the overall performance, see \cref{fig:lambda}. Note that even in such an extreme case, \ourmodel compares favorably against the best baselines (LKA and FreqFiT in \cref{tab:sota}).


\begin{table}[!htbp]
\caption{\textbf{MoR \emph{versus} previous PEFT methods for \underline{unimodal} medical image classification}.
}
\label{tab:uni}
\renewcommand{\arraystretch}{1.1}
\resizebox{\linewidth}{!}
{
\begin{tabular}{@{}lr rrrr@{}}
\toprule
\textbf{Method} & \textbf{Params}.(K) & \textbf{MEAN} & \emph{MMC-AMD} & \emph{Derm7pt} & \emph{MRNet} \\
\midrule
VFM: \textbf{CLIP} \\
FPT & 86,180 & 0.574 & 0.681 & 0.315 & 0.727 \\
UORA, {\scriptsize ACL25}\cite{uora}  & 50 & 0.607 & 0.678 & 0.419 & 0.723\\
LP & 3 & 0.610 & 0.682 & 0.413 & 0.734 \\
VPT, {\scriptsize ECCV22}\cite{vpt} & 49 & 0.616 & 0.678 & 0.425 & 0.745 \\
AdaLoRA, {\scriptsize ICLR23}\cite{adalora} & 741 & 0.623  & 0.682  & 0.459  & 0.729 \\ 

LoSA, {\scriptsize CVPR24}\cite{losa} & 100 & 0.626 & 0.698 & 0.426 & 0.753 \\
RandLoRA, {\scriptsize ICLR25}\cite{randlora}  & 2,381 & 0.644  & 0.714 & 0.484 & 0.735 \\
FPT+, {\scriptsize TNNLS25}\cite{fptp}  & 910 & 0.661 & 0.702 & 0.463 & 0.819\\

LateLoRA, {\scriptsize MIDL25}\cite{latelora} & 77 & 0.689 & 0.770 & 0.504 & 0.792 \\
LoRA, {\scriptsize ICLR22}\cite{lora} & 372 & 0.700 & 0.784 & 0.510 & 0.805 \\
BitFit, {\scriptsize ACL22}\cite{bias} & 106 & 0.705 & 0.808 & 0.498 & 0.809 \\
RLRR, {\scriptsize CVPR24}\cite{rlrr} & 292 & 0.707 & 0.804 & 0.504 & 0.813 \\
SoRA, {\scriptsize EMNLP23}\cite{sora} & 741 & 0.712 & 0.791 & 0.543 & 0.801 \\
LoRAMoE, {\scriptsize ACL24}\cite{lora-moe} & 1,247 & 0.713 & 0.788 & 0.544 & 0.806 \\
LKA, {\scriptsize MICCAI25}\cite{lka}  & 612 & 0.717 & 0.800 & 0.545 & 0.806\\ 
FreqFiT, {\scriptsize MICCAI25}\cite{freqfit}  & 7,767 & 0.720 & 0.806 & 0.552 & 0.804\\
FlyLoRA, {\scriptsize NeurIPS25}\cite{flylora}  & 740 & 0.722 & 0.792 & 0.550 & 0.823\\

\rowcolor{blue!20}
MoR (\emph{conf.}) & 1,478 & 0.737 & \underline{0.820} & \underline{0.576} & 0.816 \\
\rowcolor{blue!20}
MoR & 1,339 & \textbf{0.746} & \textbf{0.821} & \textbf{0.581} & \textbf{0.837} \\
\emph{Network structure}: \\
MoR \emph{w/o} bypass & 1,293 & \underline{0.738} & 0.809 & 0.572 & 0.833 \\
PE-Router$\rightarrow$Linear & 1,478 & 0.731 & 0.807 & 0.557 & 0.831 \\
\emph{Initialization}: \\
NZ2$\rightarrow$LoRA-GA\cite{wang2024loraga} &  1,339 & 0.709 & 0.805 & 0.511 & 0.810 \\
NZ2 \emph{w/o} $Q$ & 1,339 & 0.729 & 0.803 & 0.567 & 0.818 \\
NZ2$\rightarrow$PiSSA\cite{meng2024pissa} & 1,339 & 0.724 & 0.805 & 0.532 & \underline{0.836} \\
NZ2$\rightarrow$LoRA & 1,339 & 0.730 & 0.808 & 0.570 & 0.813 \\

\midrule
VFM: \textbf{DINOv2} \\
FPT & 86,587 & 0.622 & 0.755 & 0.335 & 0.775 \\
LP & 3 & 0.665 & 0.733 & 0.518 & 0.743 \\
AdaLoRA  & 741 &  0.666 & 0.740 & 0.514 & 0.744\\ 
LoSA     & 113 & 0.666 & 0.745 & 0.511 & 0.742 \\
UORA     & 50 & 0.668 & 0.746 & 0.516 & 0.743\\
RandLoRA & 2,381 &  0.671 & 0.747 & 0.517 & 0.751\\
VPT      & 49 & 0.679 & 0.776 & 0.471 & 0.790 \\
FPT+     &  910 & 0.687 & 0.723 & 0.527 & \underline{0.812}\\
FreqFiT  & 9,956 & 0.712 & 0.785 & 0.555 & 0.797\\
LateLoRA & 77 & 0.720 & 0.782 & 0.598 & 0.779 \\
BitFit   & 106 & 0.723 & 0.803 & 0.572 & 0.794 \\
LoRAMoE  & 1,247 & 0.724 & 0.796 & 0.591 & 0.786 \\
RLRR     & 292 & 0.724 & 0.810 & 0.567 & 0.795 \\
LoRA     & 372 & 0.725 & 0.811 & 0.579 & 0.785 \\
SoRA     & 741 &  0.728 & \underline{0.830} & 0.575 & 0.778\\
FlyLoRA  & 740 &  0.736 & 0.821 & 0.581 & 0.807\\
LKA     & 612 & 0.738 & 0.808 & 0.597 & 0.808\\ 
\rowcolor{blue!20}
MoR (\emph{conf.}) & 1,478 & 0.740 & 0.823 & 0.601 & 0.795 \\
\rowcolor{blue!20}
MoR & 1,339 & \textbf{0.753} & \textbf{0.833} & \underline{0.612} & \textbf{0.814} \\
\emph{Network structure}: \\
MoR \emph{w/o} bypass & 1,293 & 0.743 & 0.822 & 0.606 & 0.802 \\
PE-Router$\rightarrow$Linear & 1,478 & 0.741 & 0.813 & 0.607 & 0.805 \\
\emph{Initialization}: \\
NZ2$\rightarrow$LoRA & 1,339 & 0.731 & 0.811 & 0.597 & 0.784 \\
NZ2 \emph{w/o} $Q$ & 1,339 & 0.738 & 0.807 & 0.607 & 0.799 \\
NZ2$\rightarrow$LoRA-GA  & 1,339 & 0.740 & \underline{0.830} & 0.596 & 0.795 \\
NZ2$\rightarrow$PiSSA  & 1,339 & \underline{0.748} & 0.823 & \textbf{0.619} & 0.803 \\

\bottomrule

\end{tabular}}
\end{table}

\subsubsection{MoR \emph{versus} Others for VFM Adaptation} \label{sssec:eval-peft}

We start with the unimodal setting\footnote{For each dataset, we train unimodal image classifiers on each modality separately and report their average performance.}. 
We compare MoR against a number of representative PEFT methods, including 11 generic (VPT \cite{vpt}, BitFit \cite{bias}, LoRA~\cite{lora}, SoRA~\cite{sora}, RLRR \cite{rlrr}, LoSA~\cite{losa}, LoRAMoE~\cite{lora-moe}, AdaLoRA~\cite{adalora}, UORA~\cite{uora}, RandLoRA \cite{randlora}, and FlyLoRA~\cite{flylora}) and 4 medical (FPT+~\cite{fptp}, LKA~\cite{lka}, FreqFiT~\cite{freqfit},  and LateLoRA~\cite{latelora}). 
%
For broader context, we include Full-Parameter Tuning (FPT) to assess the necessity of PEFT, and Linear Probing (LP) to establish a minimal-tuning baseline. This yields a total of 17 baselines.

As shown in \cref{tab:uni}, FPT and LP are the weakest, underscoring the need for advanced PEFT. Among the PEFT baselines, FlyLoRA is the best with CLIP (0.722) and LKA with DINOv2 (0.738). The proposed MoR consistently surpasses the two best baselines.

 Next, we turn to the multimodal setting. We select a subset of the above baselines that either perform relatively well in the unimodal setting (\cref{tab:uni}) or are  structurally similar to MoR: RLRR, LKA, LoRA, LoRAMoE, SoRA and FlyLoRA. Each selected baseline is used to replace MoR within the \ourmodel framework. As shown in \cref{tab:eval-mor}, MoR is again the best.
 



\begin{table}[!tbp]
\caption{\textbf{Replacing MoR with other PEFT modules in the \ourmodel framework for M3IC}. }
\label{tab:eval-mor}
\renewcommand{\arraystretch}{1.2}
\resizebox{\linewidth}{!}
{
\begin{tabular}{@{}lr lrrr@{}}
\toprule
\textbf{Method} & \textbf{Params}.(M) & \textbf{MEAN} & \emph{MMC-AMD} & \emph{Derm7pt} & \emph{MRNet} \\
\midrule
VFM: \textbf{CLIP} \\
\rowcolor{blue!20}
MoR & 8.9 & \textbf{0.835} & \textbf{0.907} & \textbf{0.729} & \textbf{0.868} \\

RLRR & 3.3 & 0.778 (-6.8\%$\downarrow$) & 0.867 & 0.628 & 0.841\\
LKA & 4.8 & 0.794 (-4.9\%$\downarrow$) & 0.872 & 0.667 & 0.842 \\ 
LoRA & 2.7 & 0.796 (-4.7\%$\downarrow$) & 0.890 & 0.643 & 0.855  \\
FlyLoRA & 5.4 & 0.796 (-4.7\%$\downarrow$) & 0.892 & 0.652 & 0.845\\
LoRAMoE & 7.8 & 0.806 (-3.5\%$\downarrow$) & 0.875 & \underline{0.682} & \underline{0.860} \\
SoRA & 5.4 & \underline{0.809} (-3.1\%$\downarrow$) & \underline{0.893} & 0.681 & 0.853\\

\midrule
VFM: \textbf{DINOv2} \\
\rowcolor{blue!20}
MoR & 8.9 & \textbf{0.854} & \textbf{0.925} & \textbf{0.786} & \textbf{0.851}\\
RLRR & 3.3 & 0.799 (-6.4\%$\downarrow$) & 0.905 & 0.668 & 0.824\\
FlyLoRA & 5.4 & 0.803 (-6.0\%$\downarrow$) & 0.902 & 0.670 & 0.836\\
LoRA & 2.7 &  0.824 (-3.5\%$\downarrow$) & 0.888 & 0.745 & 0.838\\
LoRAMoE & 7.8 &  0.831 (-2.7\%$\downarrow$) & 0.879 & \underline{0.766} & \underline{0.849}\\ 
SoRA & 5.4 & 0.832 (-2.6\%$\downarrow$) & \underline{0.923} & 0.746 & 0.827\\
LKA & 4.8 & \underline{0.833} (-2.5\%$\downarrow$) & 0.920 & 0.736 & 0.844 \\
			
\bottomrule

\end{tabular}}
\end{table}

\subsubsection{NZ2 \emph{versus} Others for MoR Initialization} \label{sssec:eval-mor-init}


We study in the unimodal setting, see \cref{tab:uni}. Replacing NZ2 with alternative initialization methods, \ie vanilla LoRA, LoRA-GA and PiSSA, results in relative performance drop of 2.1\%--5.0\% (CLIP as VFM) and 0.7\%--2.9\% (DINOv2 as VFM). Interestingly, the more sophisticated methods (PiSSA and LoRA-GA) do not consistently outperform the vanilla LoRA initialization. With CLIP as VFM, LoRA remains the best baseline. The importance of using the orthogonal matrix $Q$ to maximize the rank of $A$ is also confirmed.

\subsubsection{In-depth Analysis of MoR Routing} \label{sssec:eval-router}


Concerning the implementation of the MoR router, replacing PE-Router by a standard linear-layer-based router causes performance drops (CLIP: 0.746$\rightarrow$0.731, DINOv2: 0.753$\rightarrow$0.741), see \cref{tab:uni}. PE-Router is both more effective and more parameter-efficient.

Removing the bypass path from PE-Router also leads to consistent performance degeneration across all datasets and VFMs. This verifies the effectiveness of the bypass design.



\begin{figure}[!htbp]
  \centering
  \subfloat[MMC-AMD]{
    \includegraphics[width=0.9\columnwidth]{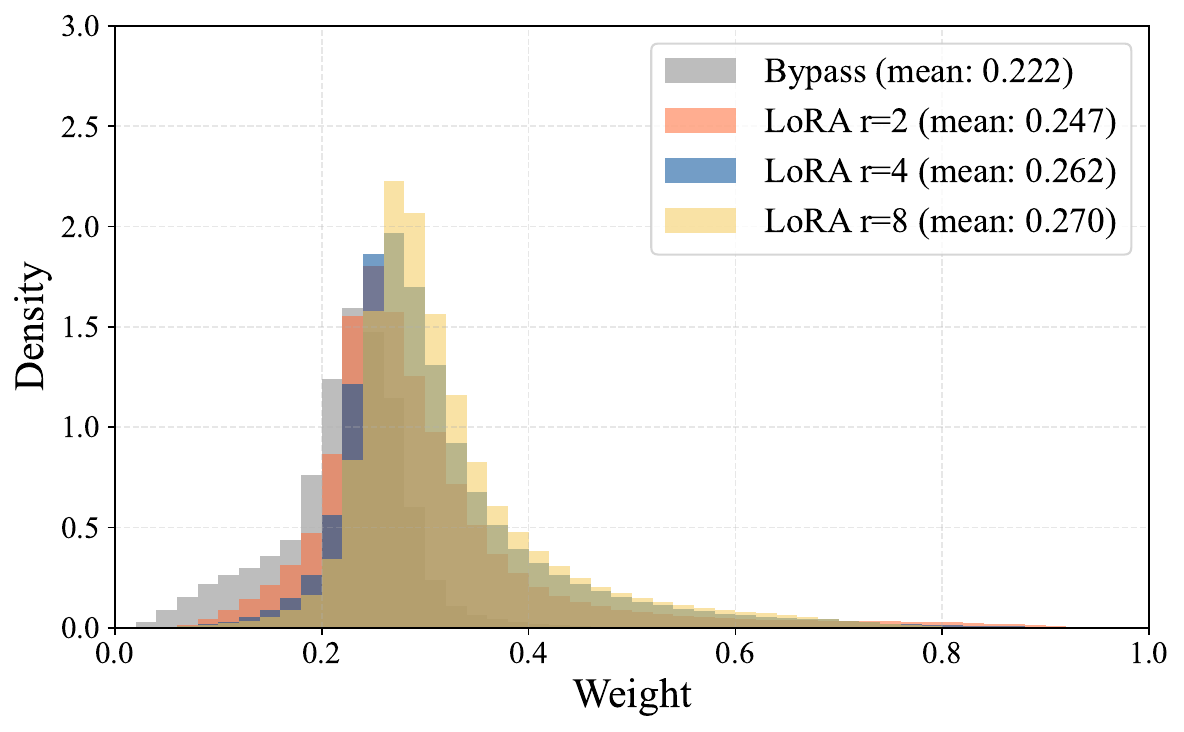}
    \label{fig:router-amd}
  }
  \vfill
  \subfloat[Derm7pt]{
    \includegraphics[width=0.9\columnwidth]{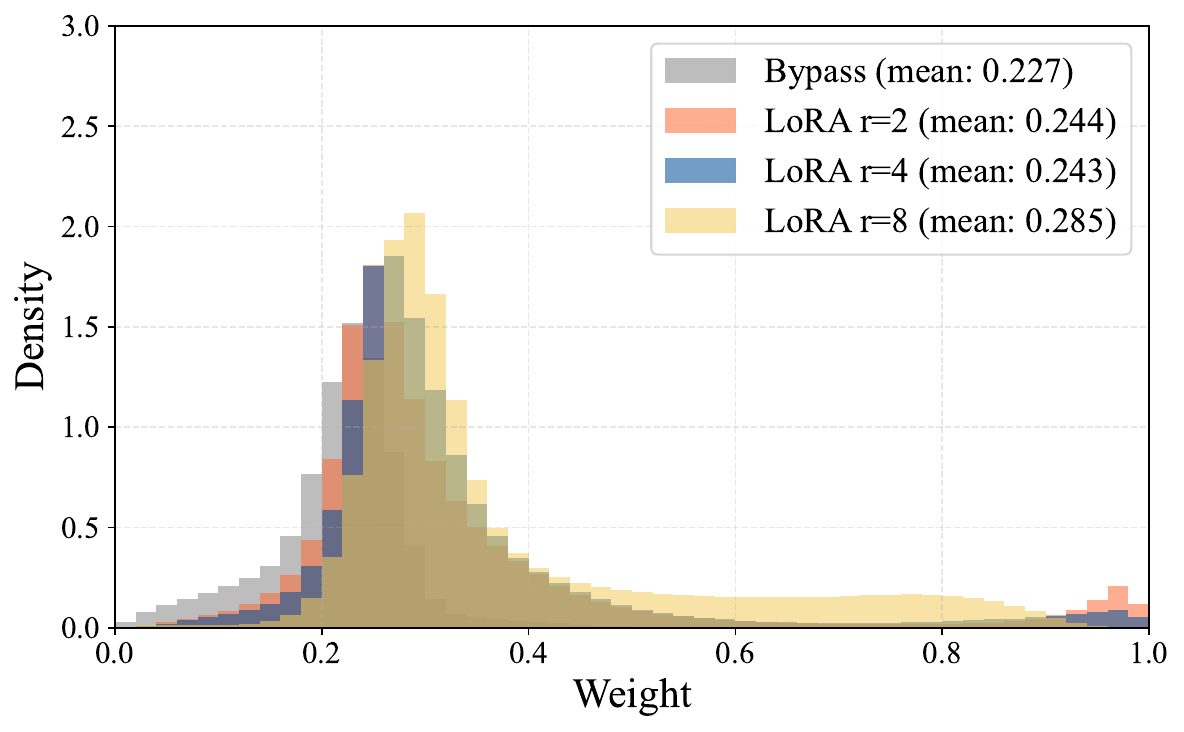}
    \label{fig:router-derm}
  }
  \vfill
  \subfloat[MRNet]{
    \includegraphics[width=0.9\columnwidth]{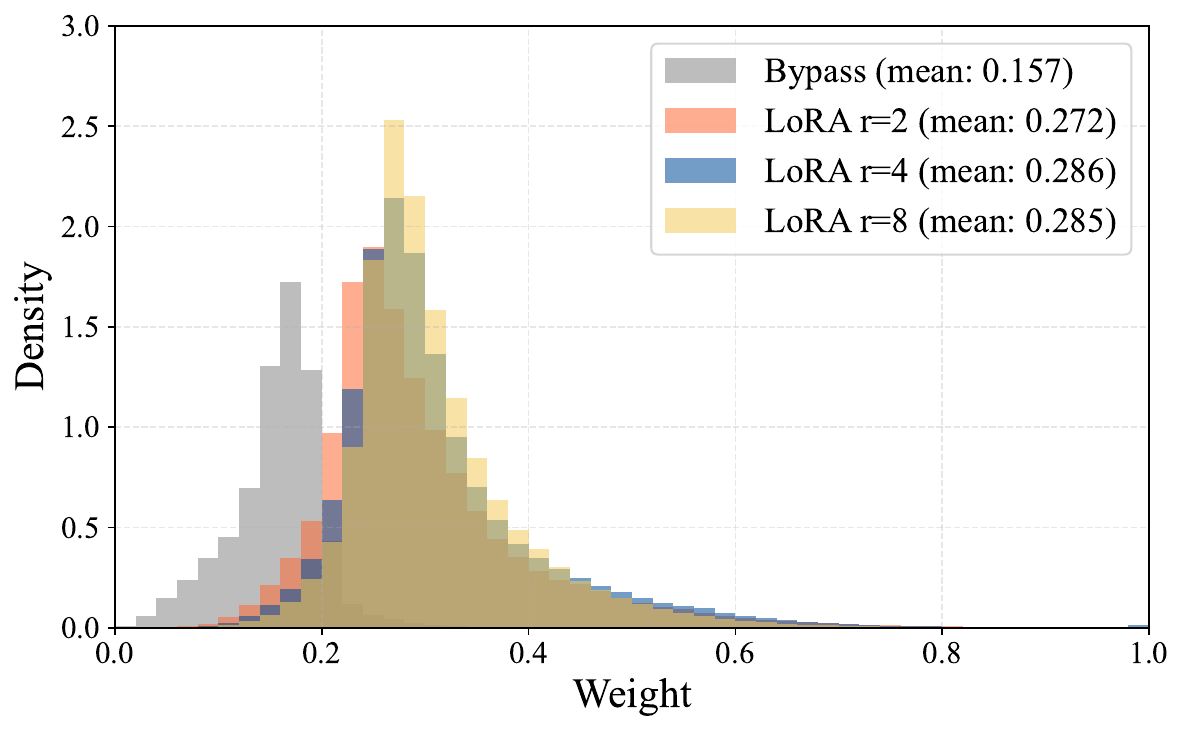}
    \label{fig:router-mrnet}
  }
  \caption{
  \textbf{Distribution of MoR router weights assigned to the three LoRA experts and the bypass path}. The weights are aggregated across all adapter layers and samples. VFM: DINOv2.
 }
  \label{fig:router-weights}
\end{figure}

\cref{fig:router-weights} presents the distribution of PE-router weights assigned to the three LoRA experts and the bypass path, with DINOv2 as VFM. Smaller bypass weights indicate stronger adaptation of the VFM. As shown in \cref{fig:router-mrnet}, the bypass weight distribution on MRNet is the most left- skewed among the three datasets. We therefore hypothesize that the domain gap between DINOv2 and MRNet may be the largest, compared to MMC-AMD and Derm7pt. To test this hypothesis, we compute the discriminability of the \emph{frozen} DINOv2 features for each dataset, defined as the ratio of the intra-class cosine distance to the inter-class counterpart. Indeed, MRNet exhibits the highest ratio (0.975), followed by Derm7pt (0.920) and MMC-AMD (0.818). These results collectively demonstrate the viability of the MoR routing for dynamically regulating the adaptation strength in an instance-specific manner.

\subsubsection{Generic \emph{or} Specialized VFMs?} \label{sssec:eval-vfm}

We compare with four domain-specific VFMs: RETFound \cite{retfound} for fundus images, PanDerm \cite{panderm} for skin images, RadioDINO \cite{radiodino} for radiological images, and BiomedCLIP \cite{bioclip} for general biomedical images. The first three are used in the \ourmodel framework for their corresponding tasks (RETFound for MMC-AMD, PanDerm for Derm7pt, RadioDINO for MRNet), while BiomedCLIP is evaluated across all three datasets. To ensure a fair comparison, we use the ViT-B/16 versions where available. An exception is RETFound, for which only a larger ViT-L/16 version is provided. As a larger VFM typically yields better performance, such model size discrepancy may confer a performance advantage to RETFound.




As shown in \cref{tab:medfm}, the four specialized VFMs are inferior to their generic counterparts. We attribute this to the richer feature representations and superior generalization capability of generic VFMs. In light of substantial resources required to develop domain-specific models, our finding suggests that effectively adapting generic VFMs is more efficient and promising. 


\begin{table}[!htbp]
\centering
\caption{\textbf{Specialized \emph{versus} generic VFMs}. Since the number of VFMs used in \ourmodel scales with the number of modalities, running on MRNet needs more VFMs and thus more parameters. 
}
\label{tab:medfm}
\renewcommand{\arraystretch}{1.2}
\resizebox{\linewidth}{!}{
\begin{tabular}{@{}lrrrr@{}}
\toprule

\textbf{VFM} & \textbf{Params. (M)} & \textbf{mAP} & \textbf{mAUC} & \textbf{mS2} \\

\midrule
\multicolumn{3}{@{}l}{Dataset: \textbf{MMC-AMD}} \\
BiomedCLIP {\scriptsize NEJMAI25}\cite{bioclip} &   6.5 & 0.775 & 0.900 & 0.772 \\
RETFound {\scriptsize Nature23}\cite{retfound}  &  15.9 & 0.811 & 0.915 & 0.835 \\ 
CLIP & 6.5 & \underline{0.907} & \underline{0.964} & \underline{0.864} \\
DINOv2 & 6.5 & \textbf{0.925} & \textbf{0.969} & \textbf{0.885} \\
\hline

\multicolumn{3}{@{}l}{Dataset: \textbf{Derm7pt}} \\
BiomedCLIP &  6.5 & 0.615 & 0.886 & 0.600 \\

PanDerm {\scriptsize NatMed25}\cite{panderm}   &  6.5 & 0.618 & 0.890 & 0.608\\

CLIP & 6.5 & \underline{0.729} & \underline{0.927} & \underline{0.779} \\

DINOv2 & 6.5 & \textbf{0.786} & \textbf{0.936} & \textbf{0.837} \\

\hline
\multicolumn{3}{@{}l}{Dataset: \textbf{MRNet}} \\
RadioDINO {\scriptsize CBM25} \cite{radiodino} & 11.6 & 0.786 & 0.774 & 0.523\\
BiomedCLIP & 11.6 & 0.811 & 0.828 & 0.666\\
DINOv2 & 11.6 & \underline{0.851} & \underline{0.855} & \underline{0.723}\\
CLIP & 11.6 & \textbf{0.868} & \textbf{0.877} & \textbf{0.738} \\
\bottomrule
\end{tabular}}
\end{table}



\section{Conclusions} \label{sec:conc}

The proposed \ourmodel framework effectively addresses two major challenges in multimodal medical image classification (M3IC): (i) the information bottleneck inherent to the delayed fusion paradigm, which limits full utilization of multimodal data, and (ii) the difficulty of applying cutting-edge VFMs to medical imagery due to the substantial domain gap. Extensive experiments on three public datasets verify the efficacy of our framework, yielding key findings as follows.

Regarding \intok token acquisition and usage, optimal performance is achieved by using the final CLS tokens together with four contextual tokens from reference modalities as the \intok tokens, and then inserting them before the first Transformer layer of the target VFM. For VFM adaptation, both the use of multiple varied-rank LoRAs and the bypass path in the router contribute to the strong performance of MoR. In particular, our router can be implemented with 75\% fewer parameters than a standard counterpart without sacrificing performance. Our NZ2 strategy further improves MoR initialization. In addition, generic VFMs, when properly adapted, outperform their domain-specific counterparts. Both the \ourmodel framework and the MoR module matter, yielding a solution that surpasses a number of competitive baselines across multiple M3IC tasks including retinal disease recognition, skin lesion recognition, and knee anomaly classification.

\medskip

\textbf{Acknowledgments}. This research was supported by National Natural Science Foundation of China (62576348) and Beijing Natural Science Foundation (L254039).

\bibliographystyle{IEEEtran}
\bibliography{ref}

\end{document}